\documentclass[sigconf]{acmart}

\renewcommand\footnotetextcopyrightpermission[1]{} 
\usepackage{subfigure}

\newcommand{\NoTwo}[1]{\textcolor{green}{#1}}

\newcounter{RNum}
\renewcommand{\theRNum}{\arabic{RNum}}
\newcommand{\Remark}{\noindent\textit{\textbf{Remark}~\refstepcounter{RNum}\textbf{\theRNum}: }}
\usepackage[switch]{lineno}
\usepackage{multirow}
\usepackage{amsmath}
\newcommand{\tsc}[1]{\textsuperscript{#1}} 
\usepackage{CJK}

\usepackage{amssymb}

\copyrightyear{2021}
\acmYear{2021}
\setcopyright{acmcopyright}\acmConference[MM '21]{Proceedings of the 29th ACM International Conference on Multimedia}{October 20--24, 2021}{Virtual Event, China}
\acmBooktitle{Proceedings of the 29th ACM International Conference on Multimedia (MM '21), October 20--24, 2021, Virtual Event, China}
\acmPrice{15.00}
\acmDOI{10.1145/3474085.3481537}
\acmISBN{978-1-4503-8651-7/21/10}

\begin{document}
\fancyhead{}
\title{ARShoe: Real-Time Augmented Reality Shoe Try-on System on Smartphones}
 

\author{
	Shan An\tsc{1,2},
	\hspace{0.1em} Guangfu Che\tsc{1},
	\hspace{0.1em} Jinghao Guo\tsc{1},
	\hspace{0.1em} Haogang Zhu\tsc{2,3\footnotemark[2]},
	\hspace{0.1em} Junjie Ye\tsc{1},
	\hspace{0.1em} Fangru Zhou\tsc{1},\\
	\hspace{0.1em} Zhaoqi Zhu\tsc{1}, 
	\hspace{0.1em} Dong Wei\tsc{1}, 
	\hspace{0.1em} Aishan Liu\tsc{2}, 
	\hspace{0.1em} Wei Zhang\tsc{1}
}

\affiliation{%
	\institution{\tsc{1}JD.COM Inc. \city{Beijing} \country{China}
		\\
		\tsc{2}State Key Lab of Software Development Environment, Beihang University\city{Beijing} \country{China} \\
		\tsc{3}Beijing Advanced Innovation Center for Big Data-Based Precision Medicine\city{Beijing} \country{China}
	}
	\institution{\{anshan,
		\hspace{0.2em}cheguangfu1,
		\hspace{0.2em}guojinghao1,
		\hspace{0.2em}yejunjie12,
		\hspace{0.2em}zhoufangru,
		\hspace{0.2em}zhuzhaoqi,
		\hspace{0.2em}weidong53,
		\hspace{0.2em}zhangwei96\}@jd.com}
	\institution{\{haogangzhu,
		\hspace{0.2em}liuaishan\}@buaa.edu.cn}
}

\begin{abstract}
  Virtual try-on technology enables users to try various fashion items using augmented reality and provides a convenient online shopping experience. However, most previous works focus on the virtual try-on for clothes while neglecting that for shoes, which is also a promising task. To this concern, this work proposes a real-time augmented reality virtual shoe try-on system for smartphones, namely ARShoe. Specifically, ARShoe adopts a novel multi-branch network to realize pose estimation and segmentation simultaneously. A solution to generate realistic 3D shoe model occlusion during the try-on process is presented. To achieve a smooth and stable try-on effect, this work further develop a novel stabilization method. Moreover, for training and evaluation, we construct the very first large-scale foot benchmark with multiple virtual shoe try-on task-related labels annotated. Exhaustive experiments on our newly constructed benchmark demonstrate the satisfying performance of ARShoe. Practical tests on common smartphones validate the real-time performance and stabilization of the proposed approach.
\end{abstract}


%



\maketitle

\begin{figure}[!t]
	\centering	
	\subfigure[]{
		\includegraphics[width=0.93\linewidth]{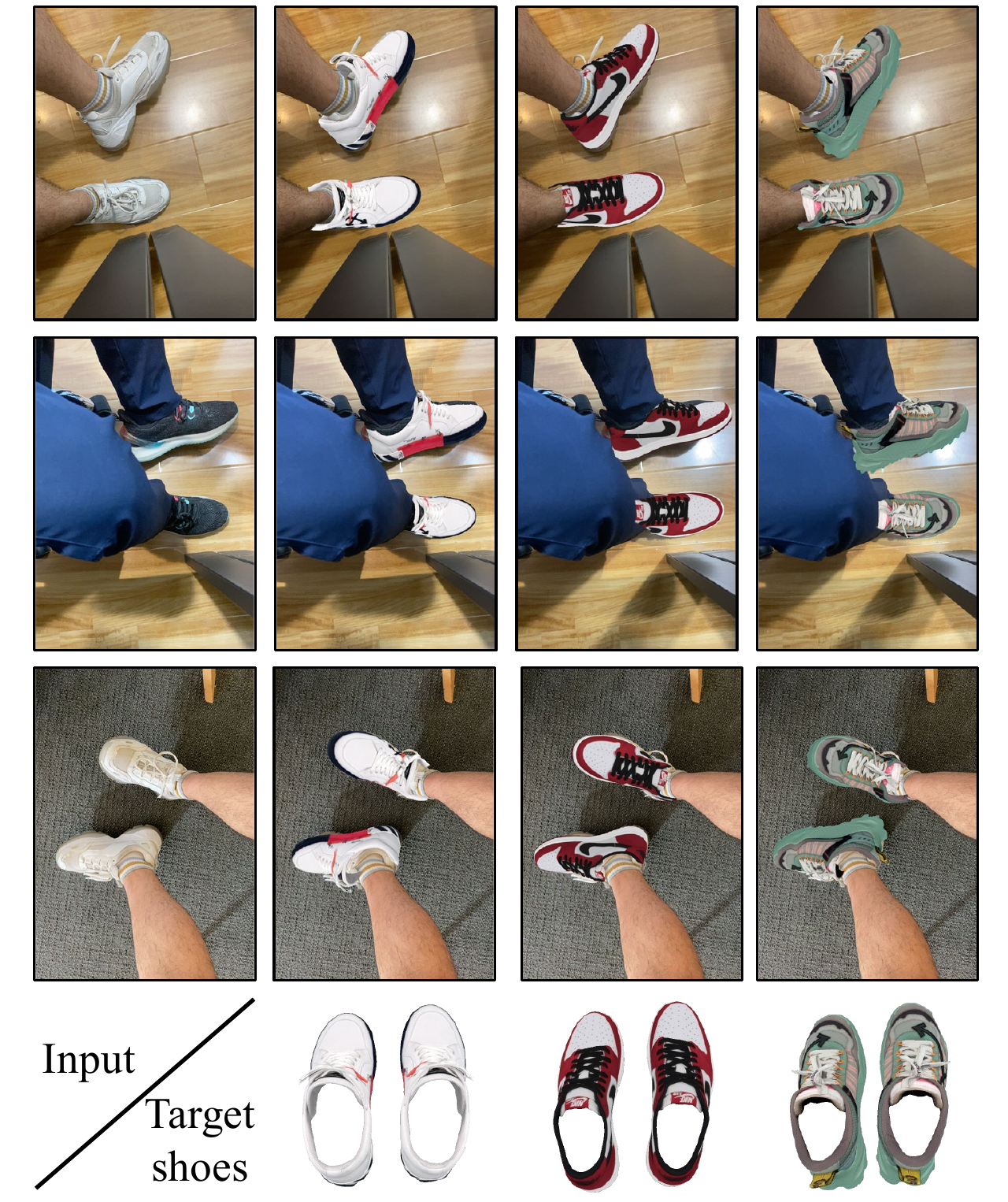} }
	\subfigure[]{
		\includegraphics[width=0.93\linewidth]{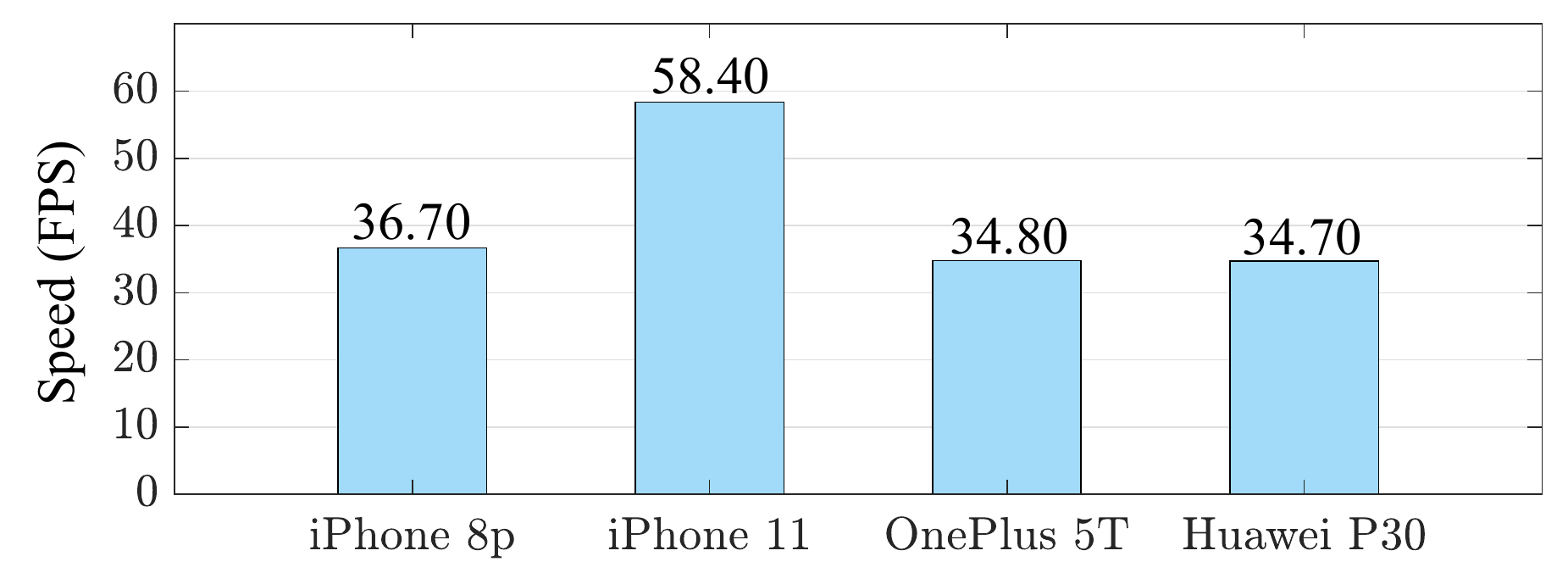}}
\setlength{\abovecaptionskip}{-1pt}
	\setlength{\belowcaptionskip}{-10pt}
	\caption{(a) Virtual shoe try-on results generated by ARShoe on a smartphone. The last row shows the target shoes, while the first column shows the original input. (b) The speed of our virtual shoe try-on system on common smartphones. ARShoe realizes a realistic virtual try-on effect with a real-time speed of over 30 frames/s (FPS) on smartphones.
	}
	\label{fig:fig1}
\end{figure}

\section{Introduction}
Recent years have witnessed the huge commercial potential of online shopping. Considerable research efforts have been put towards online shopping-related technologies, \textit{e.g.}, clothing retrieval~\cite{Jiang2016ACMMM}, product recommendation~\cite{yu2020ACMMM}, and virtual try-on~\cite{wang2020ACMMM}.
Dedicating to facilitate the online shopping experience of fashion items, virtual try-on is a promising technology, yet still challenging.

Previous works mainly focus on the virtual try-on for clothes~\cite{wang2020ACMMM, Hsieh2019ACMMM, han2018viton, Zheng2019ACMMM, wang2018toward}. In~\cite{wang2020ACMMM}, a fully convolutional graph neural network is proposed to achieve a more realistic virtual try-on effect with fine-grained details. Chia-Wei Hsieh \textit{et al.}~\cite{Hsieh2019ACMMM} propose a novel FashionOn network to tackle the body occlusion problem while preserving the detailed information of clothes. Utilizing a coarse-to-fine strategy, VITON~\cite{han2018viton} learns to synthesize a photo-realistic image of target clothes overlaid on a person in a specific pose. Comparing to clothes virtual try-on, virtual try-on for shoes has not been studied much so far while it is also a pivotal task. Chao-Te Chou \textit{et al.}~\cite{chou2018pivtons} focus on synthesizing images with shoes in the source photo replaced by target fashion shoes.

Since the main users of online shopping are consumers with smartphones, therefore smartphones are supposed to be a common platform of a virtual try-on system, which raises stringent requirements of real-time performance and computational efficiency of adopted algorithms. 
However, existing approaches commonly put great effort to optimize the realistic performance with severe computational burn introduced, while neglecting that real-time performance is also a crucial metric that influences user satisfaction.
In addition to limited computation resources, the sensing device on smartphones that can be used for virtual try-on is usually only a monocular camera. In that case, the adopted algorithm is also expected to realize 6-DoF pose estimation accurately and efficiently.
Moreover, considering the usage scenarios and the user's shooting angle, a system of virtual shoe try-on for common smartphones is more promising and needs comparing to virtual clothes try-on.


To this end, this work dedicates to study an efficient virtual shoe try-on system. In our concern, the task of virtual shoe try-on involves the following crucial subtasks: \textit{1)}~6-DoF pose estimation of feet wearing shoes. To generate 3D shoe models with the correct viewpoint, accurate 6-DoF feet pose estimation is important. Commonly, pose from 2D images is estimated by mapping 2D keypoints through the Perspective-n-Point (PnP) algorithm. \textit{2)}~2D pose estimation. Different from virtual clothes try-on, there are always two instances in the usage scenarios of virtual shoe try-on. Therefore, the 2D poses of feet are estimated to guide the accurate grouping of 2D keypoints into each foot. \textit{3)}~Occlusion generation. To produce a realistic virtual render effect, the overlays of target shoes should appear in a precise position with an appropriate scale. \textit{4)}~Stabilization. Since we target real-time virtual try-on, frequent jitter of the overlaid shoes on the screen will decrease the user experience.

In this work, we construct a neat and novel framework namely ARShoe to realize the above subtasks simultaneously and efficiently. Specifically, an efficient and effective multi-branch network is proposed to learn multi-tasks jointly, \textit{i.e.}, 2D keypoints detection, 2D pose estimation, and segmentation. To achieve realistic occlusion, we utilize the segmentation mask of the human leg, combined with the silhouette of the 3D shoe model with a transparent shoe opening, to divide the virtual occlusion area accurately. As for the stabilization operation, since we found that common stabilization and filtering methods, such as Kalman filter, and one Euro filter~\cite{casiez20121}, perform unsatisfactorily in our case. Therefore, we propose a novel method to smooth the trajectory and eliminate jitters.

Since there is no foot benchmark for virtual shoe try-on task so far, we construct and annotate a large-scale pioneer benchmark for training and evaluation. Containing a total of 86K images, all feet in our benchmark are annotated with all virtual shoe try-on-related labels, including the 8 keypoints of each foot, and the segmentation mask of legs and feet. It is worth mentioning that we neither use a multi-camera system for collecting nor the 3D bounding box for annotation. Instead, we propose to capture images of feet with a common monocular camera and annotate data in a novel way. As shown in Fig.~\ref{fig:fig1} (a), trained on our constructed benchmark, our approach achieves a satisfying performance in virtual shoe try-on. Speed evaluations on several widely used smartphones are shown in Fig.~\ref{fig:fig1} (b), the results validate that our approach can bring a smooth virtual try-on effect with a real-time speed of over 30 frames/s (FPS).

\begin{figure*}[htbp]
	\centering
		\includegraphics[width=0.99\linewidth]{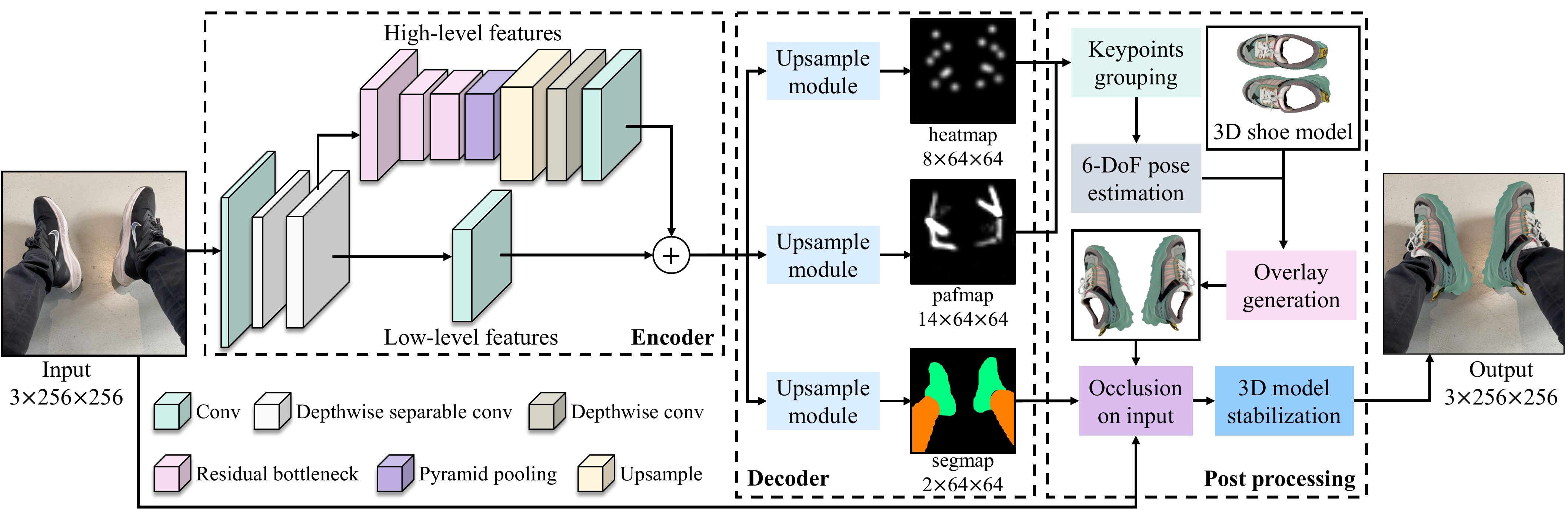}
		\\[-9pt]
	\caption{The working procedure of ARShoe. Consisting of an encoder and a decoder, the multi-branch network is trained to predict keypoints heatmap (heatmap), PAFs heatmap (pafmap), and segmentation results (segmap) simultaneously. Post processes are then performed for a smooth and realistic virtual try-on.
	}
	\label{fig:network}
\end{figure*}



The main contributions of this work are four-fold:
\begin{itemize}
	\item A novel virtual shoe try-on system for smartphones is proposed, which realizes a stable and smooth try-on effect. To our best knowledge, this is the first attempt in the literature to build a practical real-time virtual shoe try-on system. 
	\item A multi-branch network is proposed to realize keypoints detection, 2D pose estimation, and segmentation simultaneously. An origin stabilization method is presented to smooth the trajectory of overlaid shoes and eliminate jitters.
	\item A large-scale pioneer benchmark containing annotation information of keypoints and segmentation masks is constructed. An origin 6-DoF foot pose annotation method is also presented for efficient annotating.
	\item Numerous qualitative and quantitative evaluations demonstrate the satisfying realistic effect and high stability of our system. Practical tests on mainstream smartphones validate the high fluency and computational efficiency of ARShoe.  
\end{itemize}

\section{Related Work}
Note that since few studies focus on virtual shoe try-on so far, we attempt to review the most relevant works here.

\subsection{2D Pose Estimation}
When applied to the human body, 2D pose estimation refers to the joint coordinates localization, for example, 16 body joints in MPII dataset~\cite{andriluka14cvpr} and 32 body joints in Human3.6M dataset~\cite{h36m_pami}.
State-of-the-art (SOTA) 2D pose estimation approaches can be categroied into two types, \textit{i.e.}, heatmap prediction-based and coordinate regression-based methods.
Standing as a representative approach of heatmap prediction-based sort, stacked hourglass networks~\cite{newell2016stacked} utilizes symmetric hourglass-like network structures to significantly improve the accuracy of pose estimation.
Coordinate regression-based methods directly regress the joint coordinates, which has been less studied in recent years owing to lacking spatial and contextual information~\cite{fan2015combining, carreira2016human}.
There are also related studies on hand pose estimation. For instance, nonparametric structure regularization machine (NSRM)~\cite{chen2020nonparametric} and rotation-invariant mixed graph model network (R-MGMN)~\cite{kong2020rotation} perform 2D hand pose estimation through a monocular RGB camera.
OpenPose~\cite{cao2018openpose} performs 2D pose estimation for multi-person and constructs a human foot keypoint dataset which annotates each foot with 3 keypoints.


\subsection{6-DoF Pose Estimation}
The common solution for 6-DoF pose estimation is in two-stage. First, predict 2D keypoints, and then estimate 6-DoF pose from the 2D-3D correspondences through the Perspective-n-Point (PnP) algorithm.
Attempting to predict the 3D poses of challenging objects, such as partial occlusion, BB8~\cite{rad2017bb8} adopts a convolutional neural network (CNN) to estimate the 3D poses based on the 2D projections of the corners of the objects' 3D bounding boxes.
In~\cite{tekin2018real}, the authors also follow a similar procedure.
Pixel-wise voting network (PVNet)~\cite{peng2019pvnet} predicts pixel-wise vectors pointing to the object keypoints and uses these vectors to vote for keypoint locations.
This method exhibits well performance for occluded and truncated objects.
Dense pose object detector (DPOD)~\cite{zakharov2019dpod} refines the initial poses estimations after a 6-DOF pose computed using 2D-3D correspondences between images and corresponding 3D models.
PVN3D~\cite{he2020pvn3d} extends 2D keypoint approaches, which detects 3D keypoints of objects and derives the 6D pose information using least-squares fitting.
In our case, one of the branches in our network is well-trained to predict 2D keypoints in feet, then 6-DoF poses can be obtained utilizing the PnP algorithm.


\subsection{Virtual Try-On Techniques}
%
In recent years, the virtual try-on community puts great effort to develop virtual try-on for clothes and achieves promising progress. The main approaches of virtual try-on can be divided into two categories, namely image-based and video-based methods.
For the former, VITON~\cite{han2018viton}, and VTNFP~\cite{yu2019vtnfp} generate a synthesized image of a person with a desired clothing item.
For the latter, FW-GAN~\cite{dong2019fw} synthesizes the video of virtual clothes try-on based on target images and poses.
Virtual try-on for other fashion items has not attracted as much research attention like that for clothes, such as virtual try-on for cosmetics, glasses, and shoes.
For glasses try-on, X. Yuang \textit{et al.}~\cite{yuan2016magic} reconstruct 3D eyeglasses from an image and synthesize the eyeglasses on the user's face in the selfie video.
In~\cite{zhang2018augmented}, the virtual eyeglasses model is superimposed on the human face according to the head pose estimation.
Chao-Te Chou \textit{et al.}~\cite{chou2018pivtons} propose to encode feet and shoes and decode them to the desired image. However, this approach works well in scenes with a simple background and can hardly generalize to practical usage conditions. Moreover, the huge computational cost makes it unsuitable for mobile devices.
In contrast, this work focuses on virtual shoe try-on and proposes a neat and novel framework, realizing appeal realistic rendering effect while running in real-time on smartphones.
\begin{figure}[!b]
	\centering
	\includegraphics[width=0.65\linewidth]{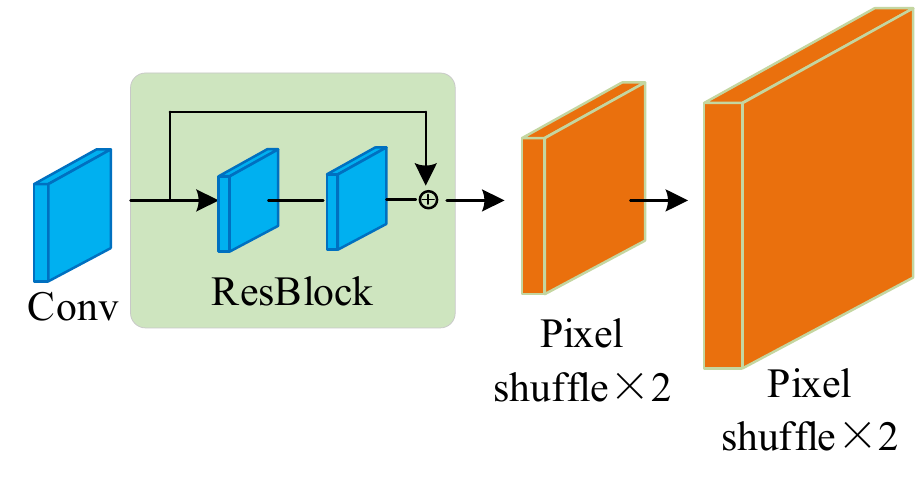}
		\\[-9pt]
	\caption{
		The architecture of the upsample module. It consists of a residual block and two consecutive pixel shuffle layers, which upsamples the $16\times16$ feature map to $64\times64$.
	}
	\label{fig:rssmodule}
\end{figure}

\section{ARShoe System}
As shown in Fig.~\ref{fig:network}, our ARShoe system has three important components: 1) A fast and accurate network for keypoints prediction, pose estimation, and segmentation; 2) A realistic occlusion generation procedure; 3) A 3D model stabilization approach. The foot pose is estimated by the predicted part affinity fields (PAFs)~\cite{cao2018openpose}, targeting to group estimated keypoints into correct foot instances. Afterward, 6-DoF poses of feet are obtained by mapping 2D keypoints from each foot. Then, the 3D shoe models can be rendered to the corresponding poses. The segmentation results are used to help find the correct area on the 3D shoe model that should be occluded by the leg. Lastly, the 3D model stabilization module ensures a smooth and realistic virtual try-on effect. Moreover, for effective and accurate annotation, an origin 6-DoF foot pose annotation method is also presented in this section. 

\subsection{Network in ARShoe}


\begin{figure}[!t]
	\centering
	\includegraphics[width=0.99\linewidth]{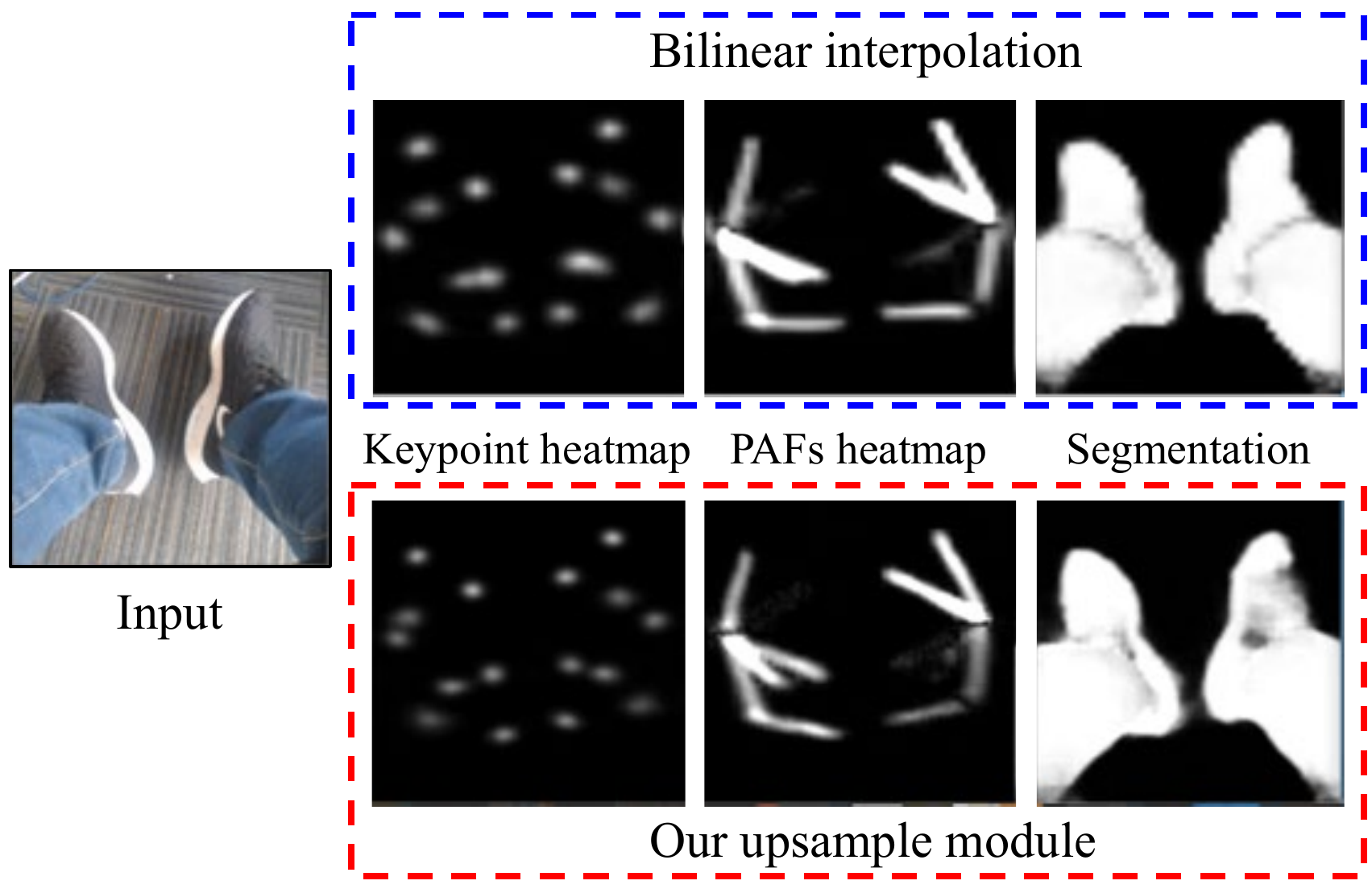}
			\\[-10pt]
	\caption{The effect of adopting different upsample methods.}
	\label{fig:rsseffect}
\end{figure}

\textbf{Network structure:}
We devise an encoder-decoder network for the 2D foot keypoints localization and the human leg and foot segmentation.
Inspired by~\cite{poudel2019fast}, the encoder network fuses high-level and low-level features for better representation.
The decoder network contains three branches: one to predict the heatmap of foot keypoints, another for PAFs prediction, and the other for human leg and foot segmentation.
For each branch, an upsampling module is applied to produce final results.

\begin{figure}[!t]
	\centering
	\includegraphics[width=0.8\linewidth]{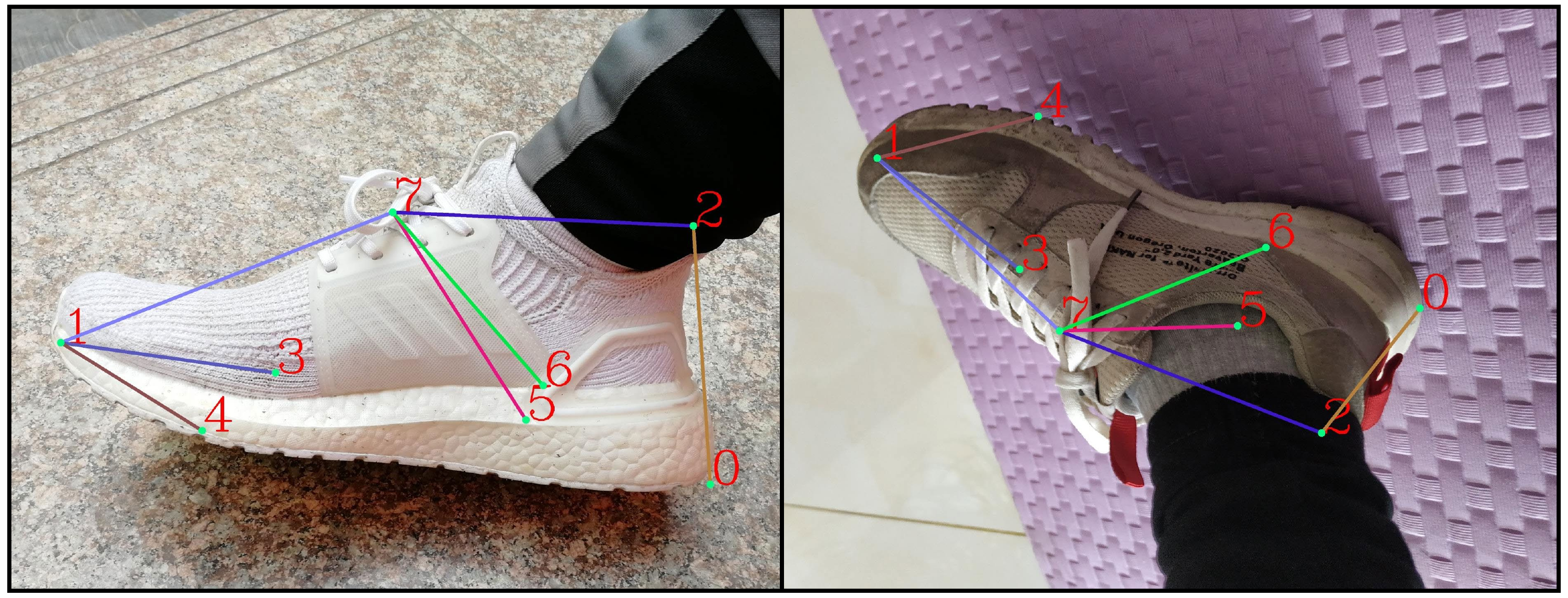}
			\\[-5pt]
	\caption{
		The illustration of the foot keypoints and their connections on the foot.
	}
	\label{fig:points}
\end{figure}

As shown in Fig.~\ref{fig:rssmodule}, each upsample module is composed of a convolution block, a ResBlock, and two pixel shuffle layers.
It can generate accurate keypoints, PAFs, and segmentation results while maintaining a relatively low computational cost.
To validate the effectiveness of our upsample module, we compare it with the commonly used upsample operation, \textit{i.e.}, bilinear interpolation operation. The results are shown in Fig.~\ref{fig:rsseffect}.
It can be seen that our upsample module generates a more accurate estimation.

In our implementation, we set 8 keypoints in the foot as shown in Fig.~\ref{fig:points}. Therefore, the output shape of the keypoints heatmap is $8\times{64}\times{64}$, corresponding to 8 keypoints respectively.
PAFs heatmap have the size of $14\times{64}\times{64}$, as show by the 7 lines in Fig.~\ref{fig:points}.
Each line has two PAFs, with the pixel value represent the projection of the line on $x$ and $y$ axis. 
The size of the segmentation results is $2\times{64}\times{64}$, with one channel for leg segmentation and another for foot segmentation. To obtain a favorable performance, we adopt $\mathcal{L}_2$ loss in all three branches.
For more details about the network architecture and the training losses, please kindly refer to supplementary material.

\begin{figure}[!t]
	\centering
	\includegraphics[width=0.85\linewidth]{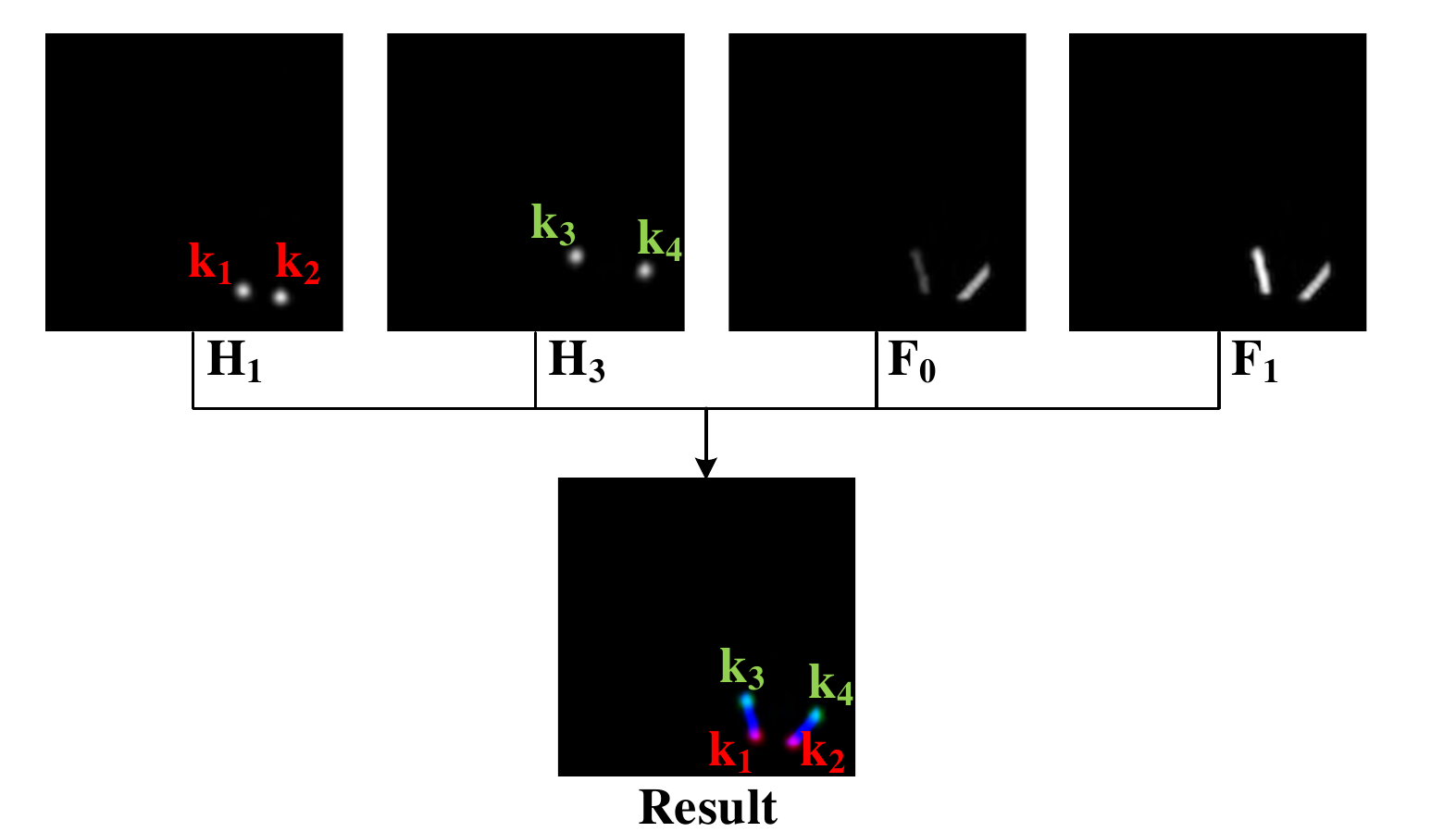} 
	\\[-10pt]
	\caption{
		Connection of adjacent keypoints.
		Two keypoints that have a connection can be grouped.
	}
	\label{fig:paf}
\end{figure}

\begin{figure}[!t]
	\centering
	\includegraphics[width=0.65\linewidth]{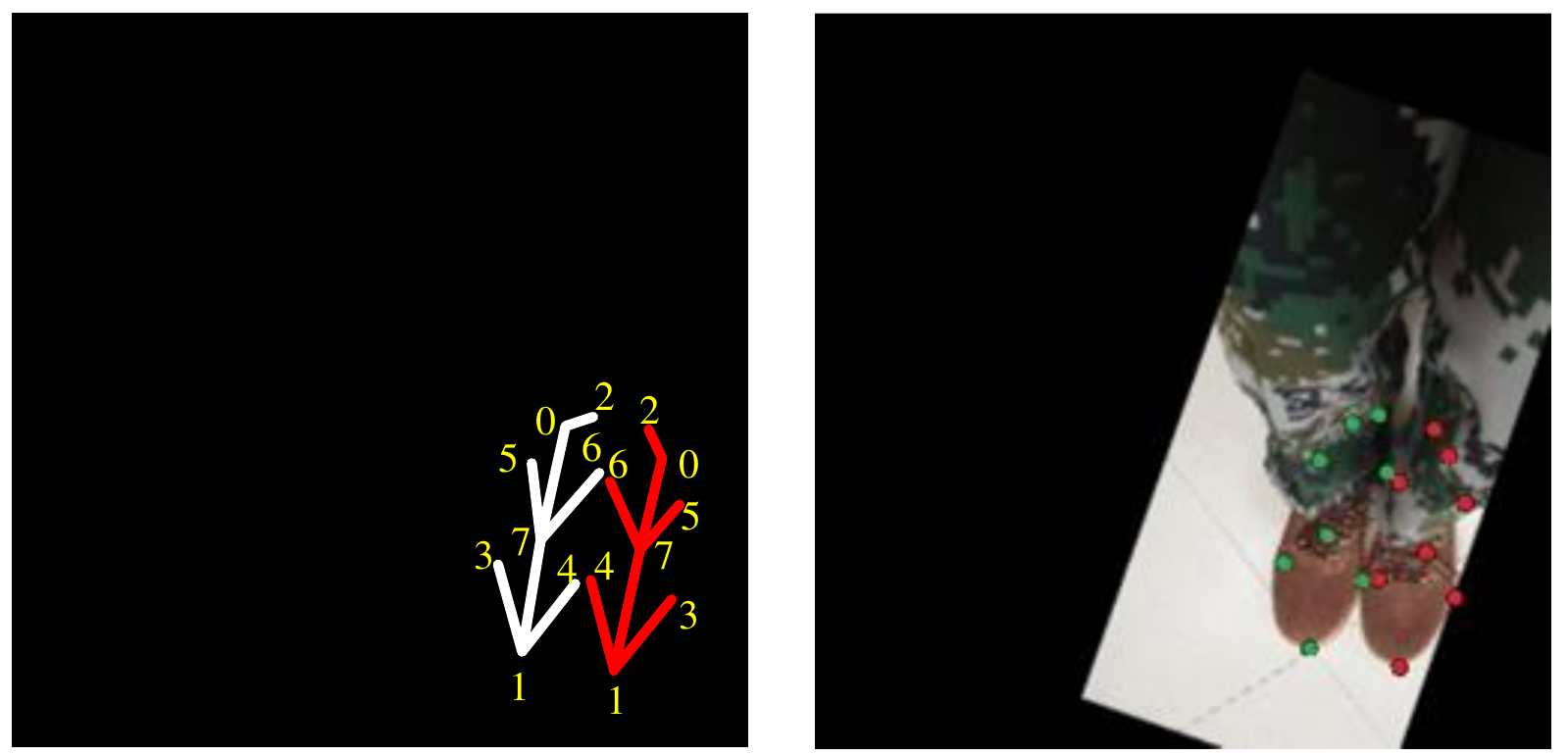}
	\caption{The grouping of all keypoints, which distinguish the two feet.
	}
	\label{fig:paf2}
\end{figure}

\textbf{Keypoints grouping:}
There are always two feet in the usage scenarios.
Therefore, we predict PAFs to represent the connection between keypoints, which can group the keypoints belonging to the same foot instance.
We derive 8 keypoint heatmaps $[H_1, ..., H_8]$ and 8 keypoint coordinates $[k_1, ..., k_8]$  from one image.
The corresponding PAFs have 7 elements, which are denoted as $[F_1, ..., F_7]$.
As shown in Fig.~\ref{fig:points}, $k_1$ and $k_3$ are connected.
The heatmap of keypoint $k_1$ and $k_3$ predicted by our network is shown in Fig.~\ref{fig:paf} as $H_1$ and $H_3$.
$F_0$ and $F_1$ represent $x$-direction and $y$-direction values of the connection relationships of keypoints $k_1$ and $k_3$ respectively, and each PAF has two connecting lines.

By looking for the two points closest to each connecting line, we can connect these two points, as shown in Fig.~\ref{fig:paf}. Keypoints $k_1$ and $k_3$ consist a connection, while keypoints $k_2$ and $k_4$ consist another.
That is, keypoints $k_1$ and $k_3$ belong to one foot, and keypoints $k_2$ and $k_4$ belong to another. In this way, all the keypoints are grouped to the corresponding foot instances. As shown in Fig.~\ref{fig:paf2}, we divided all keypoints into two groups, which means that the two feet can be distinguished in this way.

\textbf{6-DoF pose estimation:}
After keypoints grouping, each group has 8 elements, with one element represents the coordinate of one keypoint.
We use the PnP algorithm to generate the $R$ and $T$ of the human foot, according to the 3D points of a standard human foot, camera intrinsics, and the predicted 2D keypoints.

\subsection{Occlusion Generation}

To generate a realistic rendering effect, the 3D shoe model should overlay on an exact position of the input with an appropriate scale and pose. 
Prior methods rely on depth information \cite{tang2014making} or the coordinates of try-on objects in the image \cite{yuan2011mixed}.
Differently, We utilize the segmentation information of human legs, combined with the silhouette of the 3D shoe model with a transparent shoe opening, to localize the accurate virtual occlusion area.

To produce a realistic effect, we place a transparent 3D object in the area of the shoe opening.
The object position can be appropriately moved down to the inside of the shoe opening to retain a certain thickness of the edge of the shoe opening and bring a more realistic feeling.
Because the shoe opening is transparent, there will have two silhouettes after rendering. The outside silhouette is represented as $S_0$, while the inside silhouette is represented as $S_1$.

\begin{figure}[!t]
	\centering
	\includegraphics[width=0.99\linewidth]{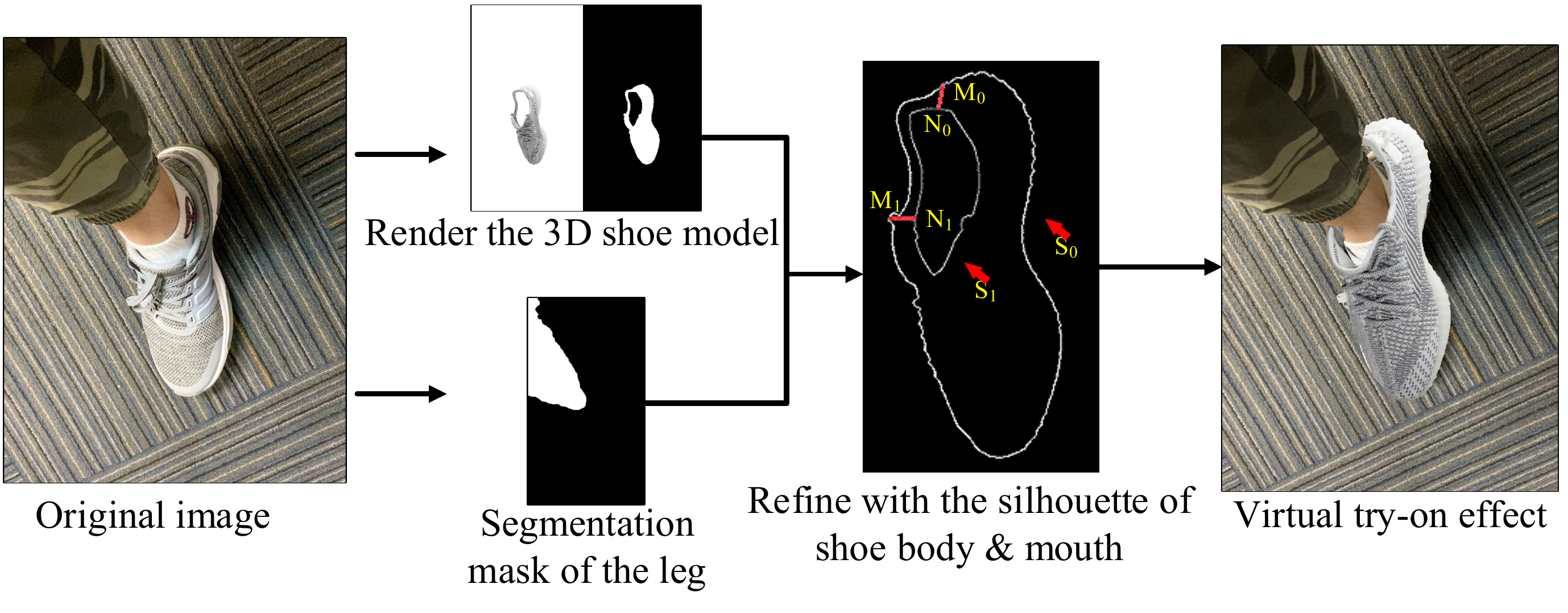}
	\caption{Occlusion generation process. According to the predicted human leg segmentation mask, the occlusion of the human leg and the 3D model is realized and the rendered image is generated.
	}
	\label{fig:occlusion}
\end{figure}

The occlusion generation process is illustrated in Fig.~\ref{fig:occlusion}.
Firstly, we find the intersections of the leg mask and $S_0$. Assuming there are two points, we denote them as $(M_0, M_1)$.
Then, we find the corresponding nearest points to $(M_0, M_1)$ in $S_1$, and represent them as $(N_0, N_1)$.
Finally, we link $(M_0, N_0)$ and $(M_1, N_1)$ to form a closed region on the mask as a 2D occlusion region and then render the region transparent for virtual occlusion.

\subsection{3D Model Stabilization}

To stabilize the overlaid 3D shoe model, we propose a method that combines the movement of corner points and the estimated 6-DoF foot pose.
As we are processing image sequence, we denote each frame as $\mathbf{I}_l$, where $l\in\{0,..., k\}$. Assuming $
\mathbf{I}_l$ is the current image, thus the previous frame is $\mathbf{I}_{l-1}$.

Firstly, we extract FAST corner points of $\mathbf{I}_{l-1}$ and $\mathbf{I}_l$ in the foot area according to the foot segmentation. These points are represented as $\mathbf{P}_{l-1}$ and $\mathbf{P}_l$. Then we use bi-direction optical flow to obtain corner points pairs $\mathbf{P}'_{l-1}=[p^1_{l-1}, ..., p^n_{l-1}]$ and $\mathbf{P}'_{l}=[p^1_{l}, ..., p^n_{l}]$, which have a matching relationship. Here $n$ is the number of matching points.

The pose of a foot is represented by a $4\times4$ Homogeneous matrix, which is from its 3D model coordinate frame to the camera coordinate frame:
\begin{equation}
	\mathbf{T}=\begin{bmatrix}
		\mathbf{R} & \mathbf{t}\\
		0 & 1
	\end{bmatrix}\in \mathbb{SE}(3) ,\quad \mathbf{R}\in \mathbb{SO}(3),\quad \mathbf{t}\in \mathbb{R}^3
\end{equation}
where $\mathbf{R}$ and $\mathbf{t} = [t^{\rm x},t^{\rm y},t^{\rm z}]^\mathsf{T}$ are the rotation matrix and the translation matrix.
The pose of the current frame is $\mathbf{T}_l$, while the pose of the previous frame is $\mathbf{T}_{l-1}$.
The point clouds of the 3D shoe model are denoted as $\mathbf{C}$, and the camera intrinsics is $\mathbf{K}$:

\begin{equation}
	\mathbf{K}= \left[
	\begin{matrix}
		f_{\rm x} & 0 & c_{\rm x} \\
		0 & f_{\rm y} & c_{\rm y} \\
		0 & 0 & 1
	\end{matrix}
	\right] \in \mathbb{R}^{3\times{3}}
	\label{eq351}
\end{equation}
where $f_{\rm x}$ and $f_{\rm y}$ are the focal length. $c_{\rm x}$ and $c_{\rm y}$ are the principal point offset.
Because of keypoint detection noise, there are inevitable jitters when using $\mathbf{T}_l$ for the virtual shoe model rendering. We leverage $\mathbf{P}'_{l-1}$ and $\mathbf{P}'_{l}$ to update a refined pose $\mathbf{T}''_l$ for the stabilization.

Firstly, we compute the average displacement $V_{\rm pix}=(v^{\rm x}_{\rm pix}, v^{\rm y}_{\rm pix})$ of corner points in pixel coordinates.
\begin{equation}
	V_{\rm pix}=\frac{1}{n}\sum_{i=1}^n(p^i_{l}-p^i_{l-1})
	\label{eq342}
\end{equation}

Secondly, we compute the translation matrix.
We assume the depths of foot in the two consecutive frames remain unchanged, \textit{i.e.}, $t^{\rm z}_l=t^{\rm z}_{l-1}$.
In camera coordinates, the average displacement ${V}_{\rm cam}$ of the corner points are:
\begin{equation}
	\begin{split}
	&v^{\rm x}_{\rm cam}=\frac{v^{\rm x}_{\rm pix} \times t^{\rm z}_l}{f_x}\\
	&v^{\rm y}_{\rm cam}=\frac{v^{\rm y}_{\rm pix} \times t^{\rm z}_l}{f_{\rm y}}\\
	&V_{\rm cam}=(v^{\rm x}_{\rm cam}, v^{\rm y}_{\rm cam}, 0)
	\end{split}
	\label{eq354}
\end{equation}

The transalation matrix is:
\begin{equation}
	\mathbf{T}'_l = [
		t^{\rm x}_l+v^{\rm x}_{\rm cam}, t^{\rm y}_l+v^{\rm y}_{\rm cam}, t^{\rm z}_l]^\mathsf{T}
	\label{eq355}
\end{equation}

We project the 3D point clouds $\mathbf{C}$ to the two image coordinate systems according to $[\mathbf{R}_{l-1}|\mathbf{T}'_l]$ and $[\mathbf{R}_l|\mathbf{T}_l]$, which results in $\mathbf{C}_1$ and $\mathbf{C}_2$:
\begin{equation}
	\begin{split}
	&\mathbf{C}_1 = \mathbf{K} \times [\mathbf{R}_{l-1}|\mathbf{T}'_l] \times \mathbf{C}\\
	&\mathbf{C}_2 = \mathbf{K} \times [\mathbf{R}_l|\mathbf{T}_l] \times \mathbf{C}
	\end{split}
	\label{eq357}
\end{equation}

The average $L1$ distance between points set $\mathbf{C}_1$ and $\mathbf{C}_2$ is:
\begin{equation}
	D = \left\Vert \mathbf{C}_1 - \mathbf{C}_2\right\Vert_1
	\label{eq358}
\end{equation}

The updating weight $w_\mathbf{R}$ of the rotation matrix is:
\begin{equation}
	w_\mathbf{R}= \alpha \times \ln(D) + \beta
	\label{eq359}
\end{equation}

Here $\alpha=0.432$ and $\beta=2.388$ are empirical parameters.
The updating weight $w_\mathbf{T}$ of the translation matrix is:
\begin{equation}
	w_\mathbf{T}= w_\mathbf{R} \times w_\mathbf{R}
	\label{eq360}
\end{equation}

Finally, we compute the updated $\mathbf{R}''_l$ and $\mathbf{T}''_l$ using the following steps:

1. The rotation matrix $\mathbf{R}_l$ and $\mathbf{R}_{l-1}$ are changed to the quaternions $\mathbf{M}_{l}$ and $\mathbf{M}_{l-1}$.

2. The quaternion $\mathbf{M}''_{l}$ is updated by:
\begin{equation}
	\mathbf{M}''_{l}= w_\mathbf{R} \times \mathbf{M}_l + (1-w_\mathbf{R}) \times \mathbf{M}_{l-1}
	\label{eq361}
\end{equation}

3. The quaternion $\mathbf{M}''_{l}$ is transformed to the rotation matrix $\mathbf{R}''_{l}$.

4. We update the translation matrix at time $l$:
\begin{equation}
	\mathbf{T}''_l= w_\mathbf{T} \times \mathbf{T}_l + (1-w_\mathbf{T}) \times \mathbf{T}'_l
	\label{eq362}
\end{equation}

Between two consecutive frames, the change of the foot pose is continuous and subtle compared with the irregular and significant noise.
To ensure the stability and consistency of the foot pose in the sequence, we first calculate the approximate position of the human foot in 3D space according to Eq.~(\ref{eq355}).
When $\mathbf{T}$ is updated, the rotation $\mathbf{t}$ contributes the most to the reprojection error.
After the weighted filtering of the rotation matrix, the change of angle will keep a certain order with the previous frame.
Hence, when the change of foot poses across two consecutive frames is slight, the pose of the previous frame will gain more weight than that of the current frame, vise versa.

\begin{figure}[t!]
	\centering
		\includegraphics[width=0.8\linewidth]{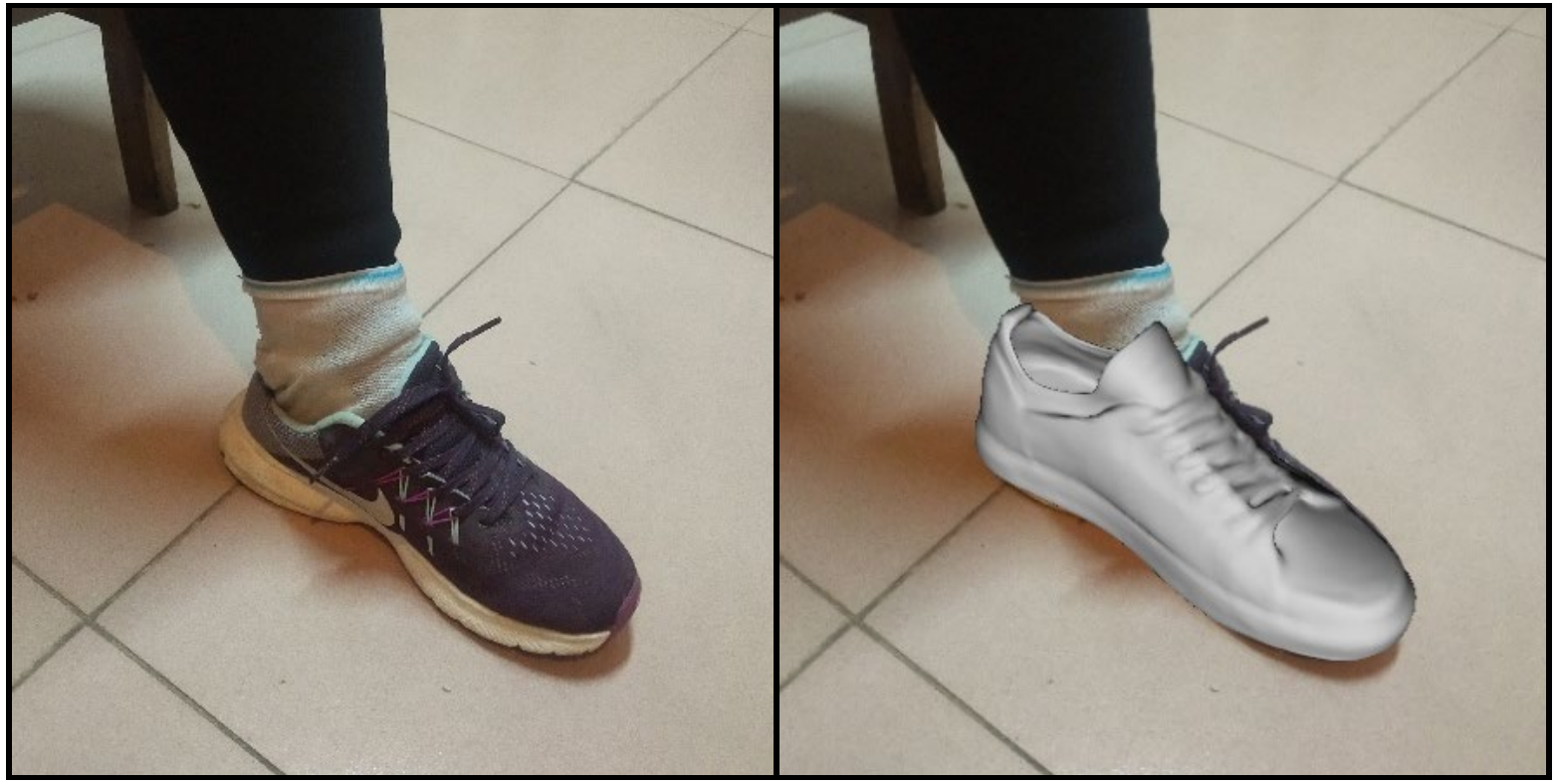}
	\caption{Illustration of the annotation. A 3D shoe model is adjusted manually to fit the shoe or foot in the image. }
	\label{fig:annotation}
\end{figure}

\subsection{6-DoF Pose Annotation }

The foot poses information is needed for training the pose estimation branch. Common 6-DoF pose annotation methods include multi-camera labeling \cite{dinesh2018carfusion, h36m_pami} and 3D bounding box labeling \cite{song2015sun, geiger2012we}. The former uses multiple cameras to shoot from different angles. According to the relationship between the cameras, the 3D information of the recovery points can be accurately labeled, but the cost is high, and the operation is complicated.
By pulling an outer bounding box, the latter realizes the selection and annotation of objects in a 2D image. However, the bounding box can be hardly labeled with high accuracy.

We designed an innovative annotation method to simulate the rendering process of the final 3D model.
Based on the 3D rendering engine, the 3D model of shoes is rotated, translated, and scaled manually to cover the position of shoes in the image, as shown in Fig.~\ref{fig:annotation}.
Therefore, the final pose transformation matrix $R$ and $T$ of the model can be obtained to achieve accurate annotation.
We selected 8 points from the rendered 3D model. The positions of these points can be seen in Fig.~\ref{fig:points}.  Keypoint 1 locates the toe. Keypoint 7 locates the center of the instep position of the foot.
Keypoints 3, 4, 5, and 6 are the points that locate the sides of the foot. Keypoints 0 and 2 are the points that locate the heel.
Seven lines connect these keypoints.
According to the transformation matrix and the camera intrinsics, the 3D points of the model can be projected to 2D point coordinates in the image, which are used as the keypoints for training.

\begin{figure}[t!]
	\centering
	\includegraphics[width=0.99\linewidth]{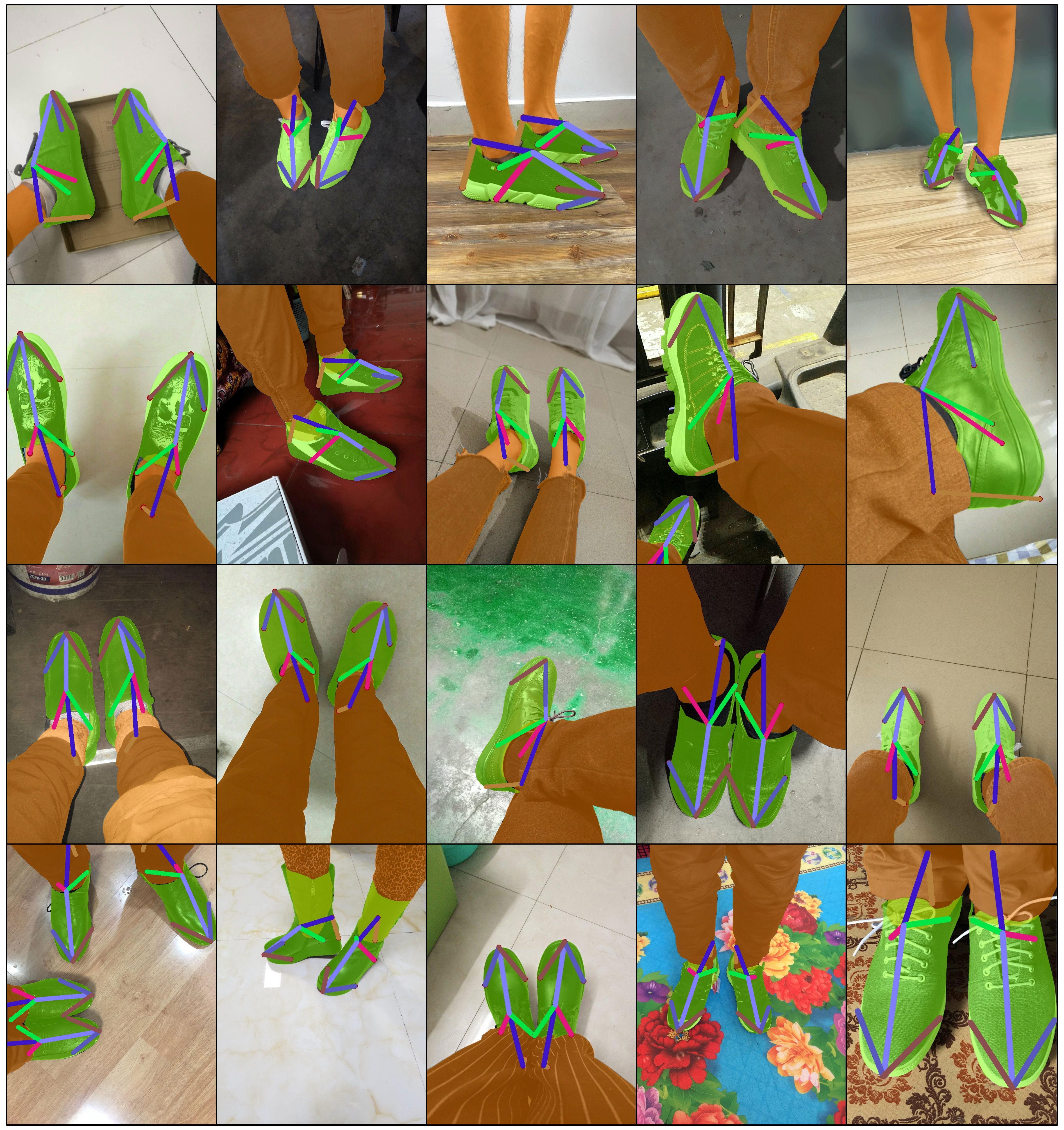}
	\caption{Some instances with annotation involved in our benchmark. The annotation information including keypoints of feet, connections of keypoints, and segmentation of feet (in \NoTwo{green}) and legs (in \textcolor{brown}{brown}).}
	\label{fig:benchmark}
\end{figure}

\section{Experiment}
In this section, exhaustive experiments are conducted to demonstrate the effectiveness of the proposed system.
Firstly, we introduce our experimental settings, including benchmark, implementation details, baselines, and evaluation metrics.
Next, we compare the proposed method with SOTA keypoint localization and segmentation approaches in terms of accuracy and speed.
Finally, we give the qualitative results to verify the immersive shoe try-on effect.

\subsection{Experimental Settings}

\textbf{Benchmark:}
This work constructs the very first large-scale foot benchmark for training and evaluation in virtual shoe try-on tasks. The benchmark contains 86,040 images of the human foot.
The annotation includes the keypoints of the foot, and the segmentation mask of human legs and feet.
Some scenes with annotation are shown in Fig.~\ref{fig:benchmark}.
The offline data augmentation is performed using rotation, translation, partial occlusion, and background changing.
The augmented dataset contains 3,012,000 images.
We use 3,000,000 images for training, and 12,000 images for testing.
All images are resized to $256\times256$.

\textbf{Implementation details:}
To train our multi-branch network, we use the Adam optimizer \cite{kingma2014adam} while setting the weight decay to $5\times10^{-4}$, and the momentum parameters $B_1 = 0.9, B_2 = 0.999$. The initial learning rate is set to $1\times10^{-4}$.
In order to speed up the model training, we use distributed training method on PyTorch 1.4 with 4 NVIDIA Tesla P40 GPUs (24GB), and the total batch size is set to 1024.
During the training process, we decrease the learning rate by 10\% at epochs 20, 30, and 60, respectively,
The entire network is trained for a total of 120 epochs.


\textbf{Baselines:}
To evaluate the performance of ARShoe in different subtasks, we compare its performance of pose estimation and foot segmentation with other SOTA approaches.
In our shoe try-on pipeline, the keypoint localization is similar to human pose estimation.
Therefore, we compare it with the SOTA human pose estimation methods, including Lightweight OpenPose~\cite{osokin2018lightweight_openpose} and Fast Pose Distillation (FPD)~\cite{zhang2019fast}.
For the segmentation performance, we compare our method with YOLACT~\cite{bolya2019yolact} and SOLOv2~\cite{wang2020solov2}.
Networks adopting different backbones are evaluated, \textit{i.e.,} Darknet\footnote{\url{https://pjreddie.com/darknet/}}, Resnet~\cite{he2016deep}, and DCN~\cite{dai2017deformable} with FPN ~\cite{lin2017feature}.

\Remark There is no open-source project about virtual shoe try-on to our best knowledge, thus comparisons are performed among ARShoe and SOTA approaches of corresponding subtasks.

\textbf{Evaluation metrics:}
For the foot keypoints localization, we use the mean average precision (mAP) at a number of object keypoint similarity (OKS) thresholds ranging from 0.5 to 0.95.
For the segmentation evaluation, we report the average precision (AP) and mean intersection of union (mIoU).

\begin{table}[!b]
	\caption{\label{tab:speed} {Comparison of computation cost and speed.}}	
	\begin{center}
		\renewcommand\tabcolsep{2pt}
		\resizebox{0.9\linewidth}{!}{
			\begin{tabular}{ccccc}
				\toprule
				\multirow{2}[0]{*}{\textbf{Method}} & \textbf{Flops}  & \textbf{Param}  & \textbf{GPU time} & \textbf{CPU time} \\
				& \textbf{(G)}    & \textbf{(M)}    & \textbf{(ms)}   & \textbf{(ms)} \\
				\midrule
				Lightweight OpenPose 	 & 4.327		& 4.092  & 14.668  & 79.433    \\
				Fast-human-pose-teacher & 20.988				& 63.595  & 64.960 & 212.880   \\
				Fast-human-pose-student & 10.244 	& 28.536  & 61.204				& 162.264  \\
				Fast-human-pose-student-FPD & 10.244 	& 28.536  & 61.204				& 162.264  \\
				\midrule	
				YOLACT-Darknet53 & 16.416 &	47.763	 & 29.660 &	1035.180 	 \\
				YOLACT-Resnet50 & 12.579 &	31.144	& 26.595 &	1116.825  \\
				YOLACT-Resnet101 & 17.441 & 	50.137 &	40.234	 & 1876.076  	 \\
				SOLOv2-Resnet50-FPN & 6.161 &	26.412 &	23.313 &	893.511  	 \\
				SOLOv2-Resnet101-FPN  & 10.081	& 45.788	 &35.411	& 1041.213  \\
				SOLOv2-DCN101-FPN  & 14.334	& 55.311 &	40.211 &	1322.024  	 \\
				\midrule
				ARShoe & \textbf{0.984} & \textbf{1.292}	&\textbf{11.844}&	\textbf{59.919}   \\
				\bottomrule
		\end{tabular}}
	\end{center}
\end{table}


\begin{table}[!b]
	\centering
	\caption{Speed analysis of ARShoe on different smartphones (ms).}
	\renewcommand\tabcolsep{2pt}
	\resizebox{1\linewidth}{!}{
		\begin{tabular}{lcccc}
			\toprule
			& iPhone 8p & iPhone 11 & OnePlus 5T & Huawei P30 \\
			\midrule
			Network forwarding & 27.278  & 4.338  & 24.037  & 25.994  \\
			Pose estimation and stablization & 13.715  & 11.226  & 28.729  & 21.578  \\
			Rendering and occlusion generation & 21.636  & 17.122  & 24.326  & 28.813  \\
			\midrule
			Overall speed & 27.278  & 17.122  & 28.729  & 28.813  \\
			\bottomrule
	\end{tabular}}
	\label{tab:phonespeed}
\end{table}%

\subsection{Computation Cost and Speed Evaluation}
The amount of calculation, model parameters, and processing speed are evaluated exhaustively to demonstrate the efficiency of the proposed system.
The involved experimental platforms include an NVIDIA Tesla P40 GPU, an Intel Xeon CPU E5-2640 v3 (2.60GHz) CPU, and different types of smartphones.

\textbf{Amount of calculation and parameters:}
We compare our network with SOTA pose estimation methods and segmentation methods in terms of calculation and parameters.
As shown in Table~\ref{tab:speed}, ARShoe has the fewest Flops (0.984 G) and parameters (1.292 M), while the second lightest approach Lightweight OpenPose still holds the Flops of 4.327 G and the parameters of 4.092 M.
For segmentation methods, SOLOv2-Resnet50-FPN is the second lightest approach, which still has 6.3 times more Flops and 20.4 times more parameters than ARShoe.
One can draw that the amount of calculation and parameters of our network is very small compared with other SOTA methods.

\Remark This is not a fair comparison for ARShoe, since our network can deal with both pose estimation and segmentation tasks simultaneously, while other involved approaches are designed for one single task. Even so, our approach still shows much fewer computational needs and parameters.


\begin{table}[!b]
	\centering
	\caption{Comparative results of foot keypoints localization.}
	\renewcommand\tabcolsep{3pt}
	\resizebox{1\linewidth}{!}{
		    \begin{tabular}{cccccc}
			\toprule
			\textbf{Method} & \textbf{mAP} & \textbf{AP$_{50}$} & \textbf{AP$_{75}$} & \textbf{AP$_{90}$} & \textbf{Speed (FPS)} \\
			\midrule
			Lightweith OpenPose~\cite{osokin2018lightweight_openpose} & 0.719  & 0.958  & 0.788  & 0.399  & 68.18 \\
			Fast-human-pose-teacher~\cite{zhang2019fast} & 0.817  & 0.976  & 0.884  & 0.617  & 15.39 \\
			Fast-human-pose-student~\cite{zhang2019fast} & 0.786  & 0.974  & 0.857  & 0.490  & 16.33 \\
			Fast-human-pose-student-FPD~\cite{zhang2019fast} & 0.797  & 0.976  & 0.863  & 0.524  & 16.33 \\
			\midrule
			ARShoe & 0.776  & 0.972  & 0.855  & 0.490  & 84.43 \\
			\bottomrule
		\end{tabular}}
	\label{tab:keypoint}%
\end{table}%


\begin{table}[htbp]
	\centering
	\caption{Comparative results of foot and leg segmentation.}
	\resizebox{0.8\linewidth}{!}{
	\begin{tabular}{ccccc}
		\toprule
		\textbf{Method} & \textbf{Backbone} & \textbf{mIoU} & \textbf{AP} & \textbf{Speed (FPS)} \\
		\midrule
		& Darknet53 & 0.895  & 0.921  & 33.72  \\
		YOLACT~\cite{bolya2019yolact} & Resnet50 & 0.890  & 0.918  & 37.60  \\
		& Resnet101 & 0.903  & 0.928  & 24.85  \\
		\midrule
		& Resnet50-FPN & 0.925  & 0.943  & 42.90  \\
		SOLOv2~\cite{wang2020solov2} & Resnet101-FPN & 0.939  & 0.952  & 28.24  \\
		& DCN101-FPN & 0.943  & 0.960  & 24.87  \\
		\midrule
		ARShoe &  None      & 0.901  & 0.927  & 84.43  \\
		\bottomrule
	\end{tabular}}
	\label{tab:seg}%
\end{table}%

\begin{figure*}[!t]
	\centering
	\includegraphics[width=0.99\linewidth]{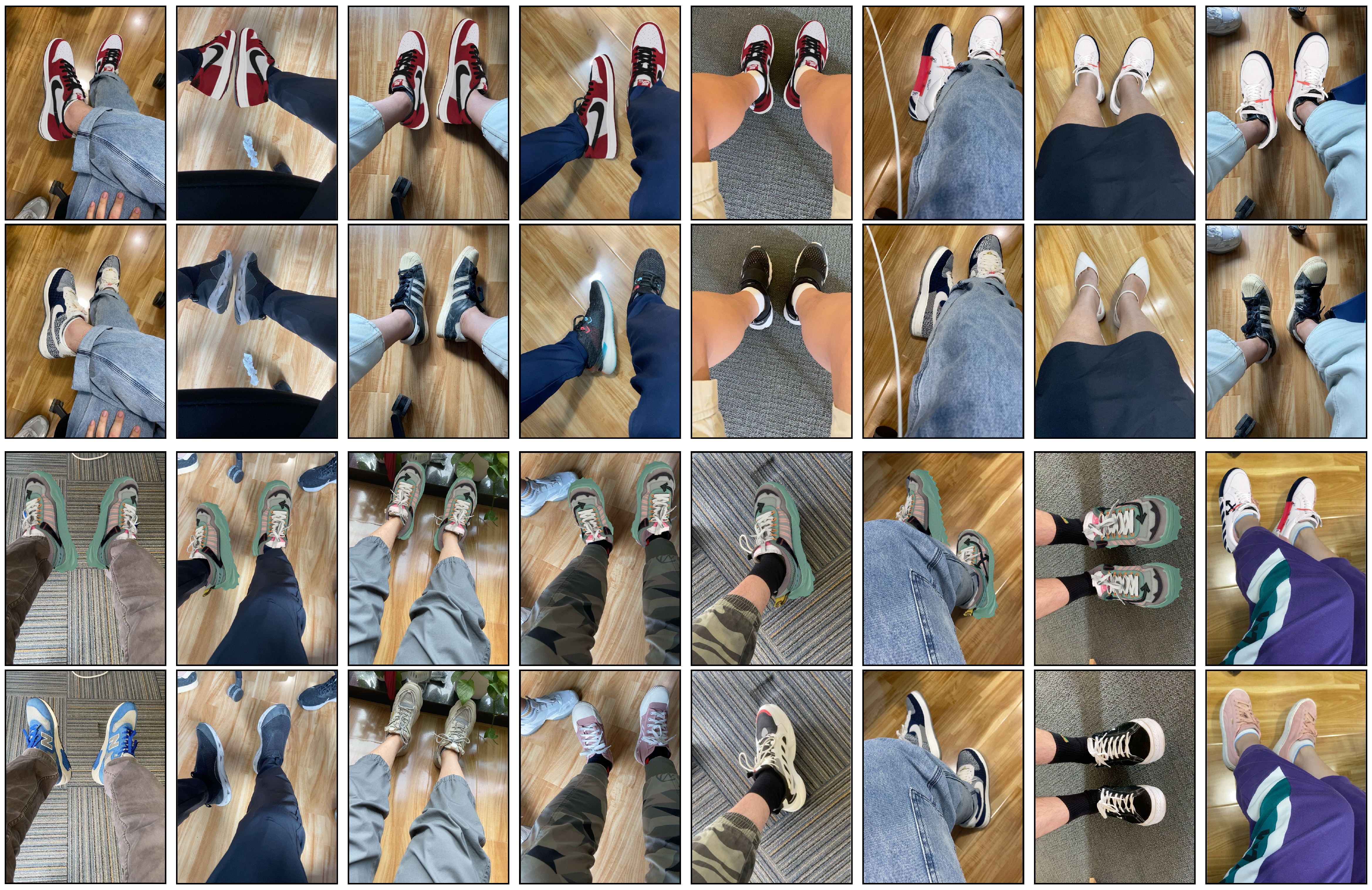}
	\caption{Virtual try-on results obtained by ARShoe. The second and fourth rows show the input images, while the first and third rows show the virtual try-on effects. ARShoe is able to handle feet in various poses and different environments, generating a realistic try-on effect. 
	}
	\label{fig:qualitative}
\end{figure*}

\textbf{Processing speed on GPU and CPU:}
The processing speed of the methods on different devices is recorded.
As shown in Table~\ref{tab:speed}, apart from appealing light-weight, our method runs efficiently on both GPU and CPU comparing to other pose estimation and segmentation approaches. Specifically, the inference time of ARShoe on a single image is 11.844 ms on GPU and 59.919 ms on CPU.


\textbf{Processing speed on smartphones:}
The above two comparisons both demonstrate the computation efficiency and light-weight of ARShoe. Therefore, we further implement the proposed approach on mobile devices to validate its practicability. Four common smartphones are adopted in this evaluation, respectively iPhone 8p, iPhone 11, OnePlus 5T, and Huawei P30. The whole architecture can be divided to three main processes, \textit{i.e.}, network forwarding, pose estimation and stabilization, and rendering and occlusion generation. Table~\ref{tab:phonespeed} reports the inference time of each processes. The speed of the whole virtual try-on system on different smartphones is also presented in Fig.~\ref{fig:fig1} (b) by FPS. We can see that ARShoe realizes a real-time speed of over 30 FPS on all four smartphones. On iPhone 11, ARShoe even reaches a speed of nearly 60 FPS. The promising results validate that ARShoe is suitable for smartphones.

\Remark Our system executes these three processes in parallel, therefore the overall frame rate is determined by the process that takes the longest time.

\subsection{Performance Evaluation}

\textbf{Keypoints localization task:}
The comparison of different keypints localization methods on the test set is shown in Table~\ref{tab:keypoint}. The GPU speed of each approach is also reported.
For a fair comparison, the involved approaches are all trained on our collected training dataset.
It can be seen that with a promising real-time speed, ARShoe achieves competitive keypoints estimation performance. Comparing to Fast-human-pose-teacher~\cite{zhang2019fast} with the best mAP, ARShoe is only 0.041 lower, while is 5.5 times faster, realizing a favorable balance of speed and accuracy. 

\textbf{Segmentation task:}
We compare our method against two SOTA segmentation methods, \textit{i.e.} YOLACT and SOLOv2, adopting different backbones.
The performance comparison and GPU speed of each method are presented in Table~\ref{tab:seg}.
Running at a fast speed of 84.43 FPS on GPU, ARShoe can still achieve a promising segmentation performance, which is not inferior to the SOTA approaches designed specifically for segmentation. 
The mIoU and AP of ARShoe are 0.901 and 0.927, which are 0.042 and 0.033 lower than the first place, respectively. 
Considering the practicability in mobile devices, our approach realizes the best balance of computation efficiency and accuracy. 

\Remark Since this work targets real-world industrial applications, we argue that the slight margin in performance is insignificant compared to the high practicability that ARShoe brings.


\textbf{6-DoF pose estimation performance:}
We also evaluate the performance of our 6-DoF pose estimation method on the test set. 
The average error of three Euler angles is 6.625$^{\circ}$, and the average distance error is only 0.384 cm.
Such satisfying 6-DoF pose estimation performance demonstrates that ARShoe can precisely handle the virtual try-on task in real-world scenarios.


\textbf{Qualitative Evaluation:}
Some virtual shoe try-on results generated by ARShoe are shown in Fig.~\ref{fig:qualitative}.
It can be seen that our system is robust to handle various conditions of different foot poses and can realize a realistic virtual try-on effect in practical scenes.
A virtual try-on video generated by ARShoe is also provided in the supplementary material.

\section{Conclusion}
This work presents a high efficient virtual shoe try-on system namely ARShoe. 
A novel multi-branch network is proposed to estimate keypoints, PAFs, and segmentation masks simultaneously and effectively. 
To realize a smooth and realistic try-on effect, an origin 3D shoe overlay generation approach and a novel stabilization method are proposed. 
Exhaustive quantitative and qualitative evaluations demonstrate the superior virtual try-on effect generated by ARShoe, with a real-time speed on smartphones.
Apart from methodology, this work constructs the very first large-scale foot benchmark with all virtual shoe try-on task-related information annotated. 
To sum up, we strongly believe both the novel ARShoe virtual shoe try-on system and the well-constructed foot benchmark will significantly contribute to the community of virtual try-on. 

\section{Acknowledgments}
This work was supported by grants from the National Key Research and Development Program of China (Grant No. 2020YFC2006200).
\bibliographystyle{ACM-Reference-Format}
\bibliography{sample-base}


\begin{thebibliography}{40}


\ifx \showCODEN    \undefined \def \showCODEN     #1{\unskip}     \fi
\ifx \showDOI      \undefined \def \showDOI       #1{#1}\fi
\ifx \showISBNx    \undefined \def \showISBNx     #1{\unskip}     \fi
\ifx \showISBNxiii \undefined \def \showISBNxiii  #1{\unskip}     \fi
\ifx \showISSN     \undefined \def \showISSN      #1{\unskip}     \fi
\ifx \showLCCN     \undefined \def \showLCCN      #1{\unskip}     \fi
\ifx \shownote     \undefined \def \shownote      #1{#1}          \fi
\ifx \showarticletitle \undefined \def \showarticletitle #1{#1}   \fi
\ifx \showURL      \undefined \def \showURL       {\relax}        \fi
\providecommand\bibfield[2]{#2}
\providecommand\bibinfo[2]{#2}
\providecommand\natexlab[1]{#1}
\providecommand\showeprint[2][]{arXiv:#2}

\bibitem[\protect\citeauthoryear{Andriluka, Pishchulin, Gehler, and
  Schiele}{Andriluka et~al\mbox{.}}{2014}]%
        {andriluka14cvpr}
\bibfield{author}{\bibinfo{person}{Mykhaylo Andriluka}, \bibinfo{person}{Leonid
  Pishchulin}, \bibinfo{person}{Peter Gehler}, {and} \bibinfo{person}{Bernt
  Schiele}.} \bibinfo{year}{2014}\natexlab{}.
\newblock \showarticletitle{2D Human Pose Estimation: New Benchmark and State
  of the Art Analysis}. In \bibinfo{booktitle}{\emph{Proceedings of the IEEE
  Conference on Computer Vision and Pattern Recognition (CVPR)}}.
  \bibinfo{pages}{3686--3693}.
\newblock
\urldef\tempurl%
\url{https://doi.org/10.1109/CVPR.2014.471}
\showDOI{\tempurl}


\bibitem[\protect\citeauthoryear{Bolya, Zhou, Xiao, and Lee}{Bolya
  et~al\mbox{.}}{2019}]%
        {bolya2019yolact}
\bibfield{author}{\bibinfo{person}{Daniel Bolya}, \bibinfo{person}{Chong Zhou},
  \bibinfo{person}{Fanyi Xiao}, {and} \bibinfo{person}{Yong~Jae Lee}.}
  \bibinfo{year}{2019}\natexlab{}.
\newblock \showarticletitle{YOLACT: Real-Time Instance Segmentation}. In
  \bibinfo{booktitle}{\emph{Proceedings of the IEEE/CVF International
  Conference on Computer Vision (ICCV)}}. \bibinfo{pages}{9156--9165}.
\newblock
\urldef\tempurl%
\url{https://doi.org/10.1109/ICCV.2019.00925}
\showDOI{\tempurl}


\bibitem[\protect\citeauthoryear{Cao, Hidalgo, Simon, Wei, and Sheikh}{Cao
  et~al\mbox{.}}{2021}]%
        {cao2018openpose}
\bibfield{author}{\bibinfo{person}{Zhe Cao}, \bibinfo{person}{Gines Hidalgo},
  \bibinfo{person}{Tomas Simon}, \bibinfo{person}{Shih-En Wei}, {and}
  \bibinfo{person}{Yaser Sheikh}.} \bibinfo{year}{2021}\natexlab{}.
\newblock \showarticletitle{OpenPose: Realtime Multi-Person 2D Pose Estimation
  Using Part Affinity Fields}.
\newblock \bibinfo{journal}{\emph{IEEE Transactions on Pattern Analysis and
  Machine Intelligence}} \bibinfo{volume}{43}, \bibinfo{number}{1}
  (\bibinfo{year}{2021}), \bibinfo{pages}{172--186}.
\newblock
\urldef\tempurl%
\url{https://doi.org/10.1109/TPAMI.2019.2929257}
\showDOI{\tempurl}


\bibitem[\protect\citeauthoryear{Carreira, Agrawal, Fragkiadaki, and
  Malik}{Carreira et~al\mbox{.}}{2016}]%
        {carreira2016human}
\bibfield{author}{\bibinfo{person}{João Carreira}, \bibinfo{person}{Pulkit
  Agrawal}, \bibinfo{person}{Katerina Fragkiadaki}, {and}
  \bibinfo{person}{Jitendra Malik}.} \bibinfo{year}{2016}\natexlab{}.
\newblock \showarticletitle{Human Pose Estimation with Iterative Error
  Feedback}. In \bibinfo{booktitle}{\emph{Proceedings of the IEEE Conference on
  Computer Vision and Pattern Recognition (CVPR)}}.
  \bibinfo{pages}{4733--4742}.
\newblock
\urldef\tempurl%
\url{https://doi.org/10.1109/CVPR.2016.512}
\showDOI{\tempurl}


\bibitem[\protect\citeauthoryear{Casiez, Roussel, and Vogel}{Casiez
  et~al\mbox{.}}{2012}]%
        {casiez20121}
\bibfield{author}{\bibinfo{person}{G\'{e}ry Casiez}, \bibinfo{person}{Nicolas
  Roussel}, {and} \bibinfo{person}{Daniel Vogel}.}
  \bibinfo{year}{2012}\natexlab{}.
\newblock \showarticletitle{1 € Filter: A Simple Speed-Based Low-Pass Filter
  for Noisy Input in Interactive Systems}. In
  \bibinfo{booktitle}{\emph{Proceedings of the SIGCHI Conference on Human
  Factors in Computing Systems (CHI)}}. \bibinfo{pages}{2527–2530}.
\newblock
\showISBNx{9781450310154}
\urldef\tempurl%
\url{https://doi.org/10.1145/2207676.2208639}
\showURL{%
\tempurl}


\bibitem[\protect\citeauthoryear{Chen, Ma, Kong, Yan, Wu, Fan, and Xie}{Chen
  et~al\mbox{.}}{2020}]%
        {chen2020nonparametric}
\bibfield{author}{\bibinfo{person}{Yifei Chen}, \bibinfo{person}{Haoyu Ma},
  \bibinfo{person}{Deying Kong}, \bibinfo{person}{Xiangyi Yan},
  \bibinfo{person}{Jianbao Wu}, \bibinfo{person}{Wei Fan}, {and}
  \bibinfo{person}{Xiaohui Xie}.} \bibinfo{year}{2020}\natexlab{}.
\newblock \showarticletitle{Nonparametric Structure Regularization Machine for
  2D Hand Pose Estimation}. In \bibinfo{booktitle}{\emph{Proceedings of the
  IEEE Winter Conference on Applications of Computer Vision (WACV)}}.
  \bibinfo{pages}{370--379}.
\newblock
\urldef\tempurl%
\url{https://doi.org/10.1109/WACV45572.2020.9093271}
\showDOI{\tempurl}


\bibitem[\protect\citeauthoryear{Chou, Lee, Zhang, Lee, and Hsu}{Chou
  et~al\mbox{.}}{2019}]%
        {chou2018pivtons}
\bibfield{author}{\bibinfo{person}{Chao-Te Chou}, \bibinfo{person}{Cheng-Han
  Lee}, \bibinfo{person}{Kaipeng Zhang}, \bibinfo{person}{Hu-Cheng Lee}, {and}
  \bibinfo{person}{Winston~H. Hsu}.} \bibinfo{year}{2019}\natexlab{}.
\newblock \showarticletitle{PIVTONS: Pose Invariant Virtual Try-On Shoe with
  Conditional Image Completion}. In \bibinfo{booktitle}{\emph{Proceedings of
  the Asian Conference on Computer Vision (ACCV)}}. \bibinfo{pages}{654--668}.
\newblock
\urldef\tempurl%
\url{https://doi.org/10.1007/978-3-030-20876-9_41}
\showDOI{\tempurl}


\bibitem[\protect\citeauthoryear{Dai, Qi, Xiong, Li, Zhang, Hu, and Wei}{Dai
  et~al\mbox{.}}{2017}]%
        {dai2017deformable}
\bibfield{author}{\bibinfo{person}{Jifeng Dai}, \bibinfo{person}{Haozhi Qi},
  \bibinfo{person}{Yuwen Xiong}, \bibinfo{person}{Yi Li},
  \bibinfo{person}{Guodong Zhang}, \bibinfo{person}{Han Hu}, {and}
  \bibinfo{person}{Yichen Wei}.} \bibinfo{year}{2017}\natexlab{}.
\newblock \showarticletitle{Deformable Convolutional Networks}. In
  \bibinfo{booktitle}{\emph{Proceedings of the IEEE International Conference on
  Computer Vision (ICCV)}}. \bibinfo{pages}{764--773}.
\newblock
\urldef\tempurl%
\url{https://doi.org/10.1109/ICCV.2017.89}
\showDOI{\tempurl}


\bibitem[\protect\citeauthoryear{Dong, Liang, Shen, Wu, Chen, and Yin}{Dong
  et~al\mbox{.}}{2019}]%
        {dong2019fw}
\bibfield{author}{\bibinfo{person}{Haoye Dong}, \bibinfo{person}{Xiaodan
  Liang}, \bibinfo{person}{Xiaohui Shen}, \bibinfo{person}{Bowen Wu},
  \bibinfo{person}{Bing-Cheng Chen}, {and} \bibinfo{person}{Jian Yin}.}
  \bibinfo{year}{2019}\natexlab{}.
\newblock \showarticletitle{FW-GAN: Flow-Navigated Warping GAN for Video
  Virtual Try-On}. In \bibinfo{booktitle}{\emph{Proceedings of the IEEE/CVF
  International Conference on Computer Vision (ICCV)}}.
  \bibinfo{pages}{1161--1170}.
\newblock
\urldef\tempurl%
\url{https://doi.org/10.1109/ICCV.2019.00125}
\showDOI{\tempurl}


\bibitem[\protect\citeauthoryear{Fan, Zheng, Lin, and Wang}{Fan
  et~al\mbox{.}}{2015}]%
        {fan2015combining}
\bibfield{author}{\bibinfo{person}{Xiaochuan Fan}, \bibinfo{person}{Kang
  Zheng}, \bibinfo{person}{Yuewei Lin}, {and} \bibinfo{person}{Song Wang}.}
  \bibinfo{year}{2015}\natexlab{}.
\newblock \showarticletitle{Combining local appearance and holistic view:
  Dual-Source Deep Neural Networks for human pose estimation}. In
  \bibinfo{booktitle}{\emph{Proceedings of the IEEE Conference on Computer
  Vision and Pattern Recognition (CVPR)}}. \bibinfo{pages}{1347--1355}.
\newblock
\urldef\tempurl%
\url{https://doi.org/10.1109/CVPR.2015.7298740}
\showDOI{\tempurl}


\bibitem[\protect\citeauthoryear{Geiger, Lenz, and Urtasun}{Geiger
  et~al\mbox{.}}{2012}]%
        {geiger2012we}
\bibfield{author}{\bibinfo{person}{Andreas Geiger}, \bibinfo{person}{Philip
  Lenz}, {and} \bibinfo{person}{Raquel Urtasun}.}
  \bibinfo{year}{2012}\natexlab{}.
\newblock \showarticletitle{Are we ready for autonomous driving? The KITTI
  vision benchmark suite}. In \bibinfo{booktitle}{\emph{Proceedings of the IEEE
  Conference on Computer Vision and Pattern Recognition (CVPR)}}.
  \bibinfo{pages}{3354--3361}.
\newblock
\urldef\tempurl%
\url{https://doi.org/10.1109/CVPR.2012.6248074}
\showDOI{\tempurl}


\bibitem[\protect\citeauthoryear{Han, Wu, Wu, Yu, and Davis}{Han
  et~al\mbox{.}}{2018}]%
        {han2018viton}
\bibfield{author}{\bibinfo{person}{Xintong Han}, \bibinfo{person}{Zuxuan Wu},
  \bibinfo{person}{Zhe Wu}, \bibinfo{person}{Ruichi Yu}, {and}
  \bibinfo{person}{Larry~S. Davis}.} \bibinfo{year}{2018}\natexlab{}.
\newblock \showarticletitle{VITON: An Image-Based Virtual Try-on Network}. In
  \bibinfo{booktitle}{\emph{Proceedings of the IEEE/CVF Conference on Computer
  Vision and Pattern Recognition (CVPR)}}. \bibinfo{pages}{7543--7552}.
\newblock
\urldef\tempurl%
\url{https://doi.org/10.1109/CVPR.2018.00787}
\showDOI{\tempurl}


\bibitem[\protect\citeauthoryear{He, Zhang, Ren, and Sun}{He
  et~al\mbox{.}}{2016}]%
        {he2016deep}
\bibfield{author}{\bibinfo{person}{Kaiming He}, \bibinfo{person}{Xiangyu
  Zhang}, \bibinfo{person}{Shaoqing Ren}, {and} \bibinfo{person}{Jian Sun}.}
  \bibinfo{year}{2016}\natexlab{}.
\newblock \showarticletitle{Deep Residual Learning for Image Recognition}. In
  \bibinfo{booktitle}{\emph{Proceedings of the IEEE Conference on Computer
  Vision and Pattern Recognition (CVPR)}}. \bibinfo{pages}{770--778}.
\newblock
\urldef\tempurl%
\url{https://doi.org/10.1109/CVPR.2016.90}
\showDOI{\tempurl}


\bibitem[\protect\citeauthoryear{He, Sun, Huang, Liu, Fan, and Sun}{He
  et~al\mbox{.}}{2020}]%
        {he2020pvn3d}
\bibfield{author}{\bibinfo{person}{Yisheng He}, \bibinfo{person}{Wei Sun},
  \bibinfo{person}{Haibin Huang}, \bibinfo{person}{Jianran Liu},
  \bibinfo{person}{Haoqiang Fan}, {and} \bibinfo{person}{Jian Sun}.}
  \bibinfo{year}{2020}\natexlab{}.
\newblock \showarticletitle{PVN3D: A Deep Point-Wise 3D Keypoints Voting
  Network for 6DoF Pose Estimation}. In \bibinfo{booktitle}{\emph{Proceedings
  of the IEEE/CVF Conference on Computer Vision and Pattern Recognition
  (CVPR)}}. \bibinfo{pages}{11629--11638}.
\newblock
\urldef\tempurl%
\url{https://doi.org/10.1109/CVPR42600.2020.01165}
\showDOI{\tempurl}


\bibitem[\protect\citeauthoryear{Hsieh, Chen, Chou, Shuai, Liu, and
  Cheng}{Hsieh et~al\mbox{.}}{2019}]%
        {Hsieh2019ACMMM}
\bibfield{author}{\bibinfo{person}{Chia-Wei Hsieh}, \bibinfo{person}{Chieh-Yun
  Chen}, \bibinfo{person}{Chien-Lung Chou}, \bibinfo{person}{Hong-Han Shuai},
  \bibinfo{person}{Jiaying Liu}, {and} \bibinfo{person}{Wen-Huang Cheng}.}
  \bibinfo{year}{2019}\natexlab{}.
\newblock \showarticletitle{FashionOn: Semantic-Guided Image-Based Virtual
  Try-on with Detailed Human and Clothing Information}. In
  \bibinfo{booktitle}{\emph{Proceedings of the ACM International Conference on
  Multimedia (MM)}}. \bibinfo{pages}{275–283}.
\newblock
\showISBNx{9781450368896}
\urldef\tempurl%
\url{https://doi.org/10.1145/3343031.3351075}
\showDOI{\tempurl}


\bibitem[\protect\citeauthoryear{Ionescu, Papava, Olaru, and
  Sminchisescu}{Ionescu et~al\mbox{.}}{2014}]%
        {h36m_pami}
\bibfield{author}{\bibinfo{person}{Catalin Ionescu}, \bibinfo{person}{Dragos
  Papava}, \bibinfo{person}{Vlad Olaru}, {and} \bibinfo{person}{Cristian
  Sminchisescu}.} \bibinfo{year}{2014}\natexlab{}.
\newblock \showarticletitle{Human3.6M: Large Scale Datasets and Predictive
  Methods for 3D Human Sensing in Natural Environments}.
\newblock \bibinfo{journal}{\emph{IEEE Transactions on Pattern Analysis and
  Machine Intelligence}} \bibinfo{volume}{36}, \bibinfo{number}{7}
  (\bibinfo{year}{2014}), \bibinfo{pages}{1325--1339}.
\newblock
\urldef\tempurl%
\url{https://doi.org/10.1109/TPAMI.2013.248}
\showDOI{\tempurl}


\bibitem[\protect\citeauthoryear{Jiang, Wu, and Fu}{Jiang
  et~al\mbox{.}}{2016}]%
        {Jiang2016ACMMM}
\bibfield{author}{\bibinfo{person}{Shuhui Jiang}, \bibinfo{person}{Yue Wu},
  {and} \bibinfo{person}{Yun Fu}.} \bibinfo{year}{2016}\natexlab{}.
\newblock \showarticletitle{Deep Bi-Directional Cross-Triplet Embedding for
  Cross-Domain Clothing Retrieval}. In \bibinfo{booktitle}{\emph{Proceedings of
  the ACM International Conference on Multimedia (MM)}}.
  \bibinfo{pages}{52–56}.
\newblock
\urldef\tempurl%
\url{https://doi.org/10.1145/2964284.2967182}
\showDOI{\tempurl}


\bibitem[\protect\citeauthoryear{Kingma and Ba}{Kingma and Ba}{2015}]%
        {kingma2014adam}
\bibfield{author}{\bibinfo{person}{Diederik~P Kingma} {and}
  \bibinfo{person}{Jimmy Ba}.} \bibinfo{year}{2015}\natexlab{}.
\newblock \showarticletitle{Adam: A Method for Stochastic Optimization}. In
  \bibinfo{booktitle}{\emph{Proceedings of the International Conference for
  Learning Representations (ICLR)}}. \bibinfo{pages}{1--15}.
\newblock


\bibitem[\protect\citeauthoryear{Kong, Ma, Chen, and Xie}{Kong
  et~al\mbox{.}}{2020}]%
        {kong2020rotation}
\bibfield{author}{\bibinfo{person}{Deying Kong}, \bibinfo{person}{Haoyu Ma},
  \bibinfo{person}{Yifei Chen}, {and} \bibinfo{person}{Xiaohui Xie}.}
  \bibinfo{year}{2020}\natexlab{}.
\newblock \showarticletitle{Rotation-invariant Mixed Graphical Model Network
  for 2D Hand Pose Estimation}. In \bibinfo{booktitle}{\emph{Proceedings of the
  IEEE Winter Conference on Applications of Computer Vision (WACV)}}.
  \bibinfo{pages}{1535--1544}.
\newblock
\urldef\tempurl%
\url{https://doi.org/10.1109/WACV45572.2020.9093638}
\showDOI{\tempurl}


\bibitem[\protect\citeauthoryear{Lin, Dollár, Girshick, He, Hariharan, and
  Belongie}{Lin et~al\mbox{.}}{2017}]%
        {lin2017feature}
\bibfield{author}{\bibinfo{person}{Tsung-Yi Lin}, \bibinfo{person}{Piotr
  Dollár}, \bibinfo{person}{Ross Girshick}, \bibinfo{person}{Kaiming He},
  \bibinfo{person}{Bharath Hariharan}, {and} \bibinfo{person}{Serge Belongie}.}
  \bibinfo{year}{2017}\natexlab{}.
\newblock \showarticletitle{Feature Pyramid Networks for Object Detection}. In
  \bibinfo{booktitle}{\emph{Proceedings of the IEEE Conference on Computer
  Vision and Pattern Recognition (CVPR)}}. \bibinfo{pages}{936--944}.
\newblock
\urldef\tempurl%
\url{https://doi.org/10.1109/CVPR.2017.106}
\showDOI{\tempurl}


\bibitem[\protect\citeauthoryear{Newell, Yang, and Deng}{Newell
  et~al\mbox{.}}{2016}]%
        {newell2016stacked}
\bibfield{author}{\bibinfo{person}{Alejandro Newell}, \bibinfo{person}{Kaiyu
  Yang}, {and} \bibinfo{person}{Jia Deng}.} \bibinfo{year}{2016}\natexlab{}.
\newblock \showarticletitle{Stacked hourglass networks for human pose
  estimation}. In \bibinfo{booktitle}{\emph{Proceedings of the European
  Conference on Computer Vision (ECCV)}}. \bibinfo{pages}{483--499}.
\newblock
\urldef\tempurl%
\url{https://doi.org/10.1007/978-3-319-46484-8_29}
\showDOI{\tempurl}


\bibitem[\protect\citeauthoryear{Osokin}{Osokin}{2018}]%
        {osokin2018lightweight_openpose}
\bibfield{author}{\bibinfo{person}{Daniil Osokin}.}
  \bibinfo{year}{2018}\natexlab{}.
\newblock \showarticletitle{Real-time 2D Multi-Person Pose Estimation on CPU:
  Lightweight OpenPose}. In \bibinfo{booktitle}{\emph{arXiv preprint
  arXiv:1811.12004}}.
\newblock


\bibitem[\protect\citeauthoryear{Peng, Liu, Huang, Zhou, and Bao}{Peng
  et~al\mbox{.}}{2019}]%
        {peng2019pvnet}
\bibfield{author}{\bibinfo{person}{Sida Peng}, \bibinfo{person}{Yuan Liu},
  \bibinfo{person}{Qixing Huang}, \bibinfo{person}{Xiaowei Zhou}, {and}
  \bibinfo{person}{Hujun Bao}.} \bibinfo{year}{2019}\natexlab{}.
\newblock \showarticletitle{PVNet: Pixel-Wise Voting Network for 6DoF Pose
  Estimation}. In \bibinfo{booktitle}{\emph{Proceedings of the IEEE/CVF
  Conference on Computer Vision and Pattern Recognition (CVPR)}}.
  \bibinfo{pages}{4556--4565}.
\newblock
\urldef\tempurl%
\url{https://doi.org/10.1109/CVPR.2019.00469}
\showDOI{\tempurl}


\bibitem[\protect\citeauthoryear{Poudel, Liwicki, and Cipolla}{Poudel
  et~al\mbox{.}}{2019}]%
        {poudel2019fast}
\bibfield{author}{\bibinfo{person}{Rudra Poudel}, \bibinfo{person}{Stephan
  Liwicki}, {and} \bibinfo{person}{Roberto Cipolla}.}
  \bibinfo{year}{2019}\natexlab{}.
\newblock \showarticletitle{Fast-SCNN: Fast Semantic Segmentation Network}. In
  \bibinfo{booktitle}{\emph{Proceedings of the British Machine Vision
  Conference (BMVC)}}. \bibinfo{pages}{1--12}.
\newblock
\urldef\tempurl%
\url{https://doi.org/10.5244/C.33.187}
\showDOI{\tempurl}


\bibitem[\protect\citeauthoryear{Rad and Lepetit}{Rad and Lepetit}{2017}]%
        {rad2017bb8}
\bibfield{author}{\bibinfo{person}{Mahdi Rad} {and} \bibinfo{person}{Vincent
  Lepetit}.} \bibinfo{year}{2017}\natexlab{}.
\newblock \showarticletitle{BB8: A Scalable, Accurate, Robust to Partial
  Occlusion Method for Predicting the 3D Poses of Challenging Objects without
  Using Depth}. In \bibinfo{booktitle}{\emph{Proceedings of the IEEE
  International Conference on Computer Vision (ICCV)}}.
  \bibinfo{pages}{3848--3856}.
\newblock
\urldef\tempurl%
\url{https://doi.org/10.1109/ICCV.2017.413}
\showDOI{\tempurl}


\bibitem[\protect\citeauthoryear{Reddy, Vo, and Narasimhan}{Reddy
  et~al\mbox{.}}{2018}]%
        {dinesh2018carfusion}
\bibfield{author}{\bibinfo{person}{N~Dinesh Reddy}, \bibinfo{person}{Minh Vo},
  {and} \bibinfo{person}{Srinivasa~G. Narasimhan}.}
  \bibinfo{year}{2018}\natexlab{}.
\newblock \showarticletitle{CarFusion: Combining Point Tracking and Part
  Detection for Dynamic 3D Reconstruction of Vehicles}. In
  \bibinfo{booktitle}{\emph{Proceedings of the IEEE/CVF Conference on Computer
  Vision and Pattern Recognition (CVPR)}}. \bibinfo{pages}{1906--1915}.
\newblock
\urldef\tempurl%
\url{https://doi.org/10.1109/CVPR.2018.00204}
\showDOI{\tempurl}


\bibitem[\protect\citeauthoryear{Song, Lichtenberg, and Xiao}{Song
  et~al\mbox{.}}{2015}]%
        {song2015sun}
\bibfield{author}{\bibinfo{person}{Shuran Song}, \bibinfo{person}{Samuel~P.
  Lichtenberg}, {and} \bibinfo{person}{Jianxiong Xiao}.}
  \bibinfo{year}{2015}\natexlab{}.
\newblock \showarticletitle{SUN RGB-D: A RGB-D scene understanding benchmark
  suite}. In \bibinfo{booktitle}{\emph{Proceedings of the IEEE Conference on
  Computer Vision and Pattern Recognition (CVPR)}}. \bibinfo{pages}{567--576}.
\newblock
\urldef\tempurl%
\url{https://doi.org/10.1109/CVPR.2015.7298655}
\showDOI{\tempurl}


\bibitem[\protect\citeauthoryear{Tang, Zhang, Tang, Xu, and Fang}{Tang
  et~al\mbox{.}}{2014}]%
        {tang2014making}
\bibfield{author}{\bibinfo{person}{Difei Tang}, \bibinfo{person}{Juyong Zhang},
  \bibinfo{person}{Ketan Tang}, \bibinfo{person}{Lingfeng Xu}, {and}
  \bibinfo{person}{Lu Fang}.} \bibinfo{year}{2014}\natexlab{}.
\newblock \showarticletitle{Making 3D Eyeglasses Try-on practical}. In
  \bibinfo{booktitle}{\emph{Proceedings of the IEEE International Conference on
  Multimedia and Expo Workshops (ICMEW)}}. \bibinfo{pages}{1--6}.
\newblock
\urldef\tempurl%
\url{https://doi.org/10.1109/ICMEW.2014.6890545}
\showDOI{\tempurl}


\bibitem[\protect\citeauthoryear{Tekin, Sinha, and Fua}{Tekin
  et~al\mbox{.}}{2018}]%
        {tekin2018real}
\bibfield{author}{\bibinfo{person}{Bugra Tekin}, \bibinfo{person}{Sudipta~N.
  Sinha}, {and} \bibinfo{person}{Pascal Fua}.} \bibinfo{year}{2018}\natexlab{}.
\newblock \showarticletitle{Real-Time Seamless Single Shot 6D Object Pose
  Prediction}. In \bibinfo{booktitle}{\emph{Proceedings of the IEEE/CVF
  Conference on Computer Vision and Pattern Recognition (CVPR)}}.
  \bibinfo{pages}{292--301}.
\newblock
\urldef\tempurl%
\url{https://doi.org/10.1109/CVPR.2018.00038}
\showDOI{\tempurl}


\bibitem[\protect\citeauthoryear{Wang, Zheng, Liang, Chen, Lin, and Yang}{Wang
  et~al\mbox{.}}{2018}]%
        {wang2018toward}
\bibfield{author}{\bibinfo{person}{Bochao Wang}, \bibinfo{person}{Huabin
  Zheng}, \bibinfo{person}{Xiaodan Liang}, \bibinfo{person}{Yimin Chen},
  \bibinfo{person}{Liang Lin}, {and} \bibinfo{person}{Meng Yang}.}
  \bibinfo{year}{2018}\natexlab{}.
\newblock \showarticletitle{Toward Characteristic-Preserving Image-Based
  Virtual Try-On Network}. In \bibinfo{booktitle}{\emph{Proceedings of the
  European Conference on Computer Vision (ECCV)}}. \bibinfo{pages}{607--623}.
\newblock
\urldef\tempurl%
\url{https://doi.org/10.1007/978-3-030-01261-8_36}
\showDOI{\tempurl}


\bibitem[\protect\citeauthoryear{Wang, Sha, Zhang, Li, and Mei}{Wang
  et~al\mbox{.}}{2020a}]%
        {wang2020ACMMM}
\bibfield{author}{\bibinfo{person}{Jiahang Wang}, \bibinfo{person}{Tong Sha},
  \bibinfo{person}{Wei Zhang}, \bibinfo{person}{Zhoujun Li}, {and}
  \bibinfo{person}{Tao Mei}.} \bibinfo{year}{2020}\natexlab{a}.
\newblock \showarticletitle{Down to the Last Detail: Virtual Try-on with
  Fine-Grained Details}. In \bibinfo{booktitle}{\emph{Proceedings of the ACM
  International Conference on Multimedia (MM)}}. \bibinfo{pages}{466–474}.
\newblock
\showISBNx{9781450379885}
\urldef\tempurl%
\url{https://doi.org/10.1145/3394171.3413514}
\showDOI{\tempurl}


\bibitem[\protect\citeauthoryear{Wang, Zhang, Kong, Li, and Shen}{Wang
  et~al\mbox{.}}{2020b}]%
        {wang2020solov2}
\bibfield{author}{\bibinfo{person}{Xinlong Wang}, \bibinfo{person}{Rufeng
  Zhang}, \bibinfo{person}{Tao Kong}, \bibinfo{person}{Lei Li}, {and}
  \bibinfo{person}{Chunhua Shen}.} \bibinfo{year}{2020}\natexlab{b}.
\newblock \showarticletitle{SOLOv2: Dynamic and Fast Instance Segmentation}. In
  \bibinfo{booktitle}{\emph{Proceedings of the Advances in Neural Information
  Processing Systems (NeurIPS)}}, Vol.~\bibinfo{volume}{33}.
  \bibinfo{pages}{17721--17732}.
\newblock


\bibitem[\protect\citeauthoryear{Yu, Wang, and Xie}{Yu et~al\mbox{.}}{2019}]%
        {yu2019vtnfp}
\bibfield{author}{\bibinfo{person}{Ruiyun Yu}, \bibinfo{person}{Xiaoqi Wang},
  {and} \bibinfo{person}{Xiaohui Xie}.} \bibinfo{year}{2019}\natexlab{}.
\newblock \showarticletitle{VTNFP: An Image-Based Virtual Try-On Network With
  Body and Clothing Feature Preservation}. In
  \bibinfo{booktitle}{\emph{Proceedings of the IEEE/CVF International
  Conference on Computer Vision (ICCV)}}. \bibinfo{pages}{10510--10519}.
\newblock
\urldef\tempurl%
\url{https://doi.org/10.1109/ICCV.2019.01061}
\showDOI{\tempurl}


\bibitem[\protect\citeauthoryear{Yu, Gan, Wei, Cheng, and Nie}{Yu
  et~al\mbox{.}}{2020}]%
        {yu2020ACMMM}
\bibfield{author}{\bibinfo{person}{Xuzheng Yu}, \bibinfo{person}{Tian Gan},
  \bibinfo{person}{Yinwei Wei}, \bibinfo{person}{Zhiyong Cheng}, {and}
  \bibinfo{person}{Liqiang Nie}.} \bibinfo{year}{2020}\natexlab{}.
\newblock \showarticletitle{Personalized Item Recommendation for Second-Hand
  Trading Platform}. In \bibinfo{booktitle}{\emph{Proceedings of the ACM
  International Conference on Multimedia (MM)}}. \bibinfo{pages}{3478–3486}.
\newblock
\urldef\tempurl%
\url{https://doi.org/10.1145/3394171.3413640}
\showDOI{\tempurl}


\bibitem[\protect\citeauthoryear{Yuan, Khan, Farbiz, Niswar, and Huang}{Yuan
  et~al\mbox{.}}{2011}]%
        {yuan2011mixed}
\bibfield{author}{\bibinfo{person}{Miaolong Yuan},
  \bibinfo{person}{Ishtiaq~Rasool Khan}, \bibinfo{person}{Farzam Farbiz},
  \bibinfo{person}{Arthur Niswar}, {and} \bibinfo{person}{Zhiyong Huang}.}
  \bibinfo{year}{2011}\natexlab{}.
\newblock \showarticletitle{A Mixed Reality System for Virtual Glasses Try-On}.
  In \bibinfo{booktitle}{\emph{Proceedings of the International Conference on
  Virtual Reality Continuum and Its Applications in Industry (VRCAI)}}.
  \bibinfo{pages}{363–366}.
\newblock
\urldef\tempurl%
\url{https://doi.org/10.1145/2087756.2087816}
\showDOI{\tempurl}


\bibitem[\protect\citeauthoryear{Yuan, Tang, Liu, Ling, and Fang}{Yuan
  et~al\mbox{.}}{2017}]%
        {yuan2016magic}
\bibfield{author}{\bibinfo{person}{Xiaoyun Yuan}, \bibinfo{person}{Difei Tang},
  \bibinfo{person}{Yebin Liu}, \bibinfo{person}{Qing Ling}, {and}
  \bibinfo{person}{Lu Fang}.} \bibinfo{year}{2017}\natexlab{}.
\newblock \showarticletitle{Magic Glasses: From 2D to 3D}.
\newblock \bibinfo{journal}{\emph{IEEE Transactions on Circuits and Systems for
  Video Technology}} \bibinfo{volume}{27}, \bibinfo{number}{4}
  (\bibinfo{year}{2017}), \bibinfo{pages}{843--854}.
\newblock
\urldef\tempurl%
\url{https://doi.org/10.1109/TCSVT.2016.2556439}
\showDOI{\tempurl}


\bibitem[\protect\citeauthoryear{Zakharov, Shugurov, and Ilic}{Zakharov
  et~al\mbox{.}}{2019}]%
        {zakharov2019dpod}
\bibfield{author}{\bibinfo{person}{Sergey Zakharov}, \bibinfo{person}{Ivan
  Shugurov}, {and} \bibinfo{person}{Slobodan Ilic}.}
  \bibinfo{year}{2019}\natexlab{}.
\newblock \showarticletitle{DPOD: 6D Pose Object Detector and Refiner}. In
  \bibinfo{booktitle}{\emph{Proceedings of the IEEE/CVF International
  Conference on Computer Vision (ICCV)}}. \bibinfo{pages}{1941--1950}.
\newblock
\urldef\tempurl%
\url{https://doi.org/10.1109/ICCV.2019.00203}
\showDOI{\tempurl}


\bibitem[\protect\citeauthoryear{Zhang}{Zhang}{2018}]%
        {zhang2018augmented}
\bibfield{author}{\bibinfo{person}{Boping Zhang}.}
  \bibinfo{year}{2018}\natexlab{}.
\newblock \showarticletitle{Augmented reality virtual glasses try-on technology
  based on iOS platform}.
\newblock \bibinfo{journal}{\emph{EURASIP Journal on Image and Video
  Processing}} \bibinfo{volume}{132}, \bibinfo{number}{2018}
  (\bibinfo{year}{2018}), \bibinfo{pages}{1--19}.
\newblock
\urldef\tempurl%
\url{https://doi.org/10.1186/s13640-018-0373-8}
\showDOI{\tempurl}


\bibitem[\protect\citeauthoryear{Zhang, Zhu, and Ye}{Zhang
  et~al\mbox{.}}{2019}]%
        {zhang2019fast}
\bibfield{author}{\bibinfo{person}{Feng Zhang}, \bibinfo{person}{Xiatian Zhu},
  {and} \bibinfo{person}{Mao Ye}.} \bibinfo{year}{2019}\natexlab{}.
\newblock \showarticletitle{Fast Human Pose Estimation}. In
  \bibinfo{booktitle}{\emph{Proceedings of the IEEE/CVF Conference on Computer
  Vision and Pattern Recognition (CVPR)}}. \bibinfo{pages}{3512--3521}.
\newblock
\urldef\tempurl%
\url{https://doi.org/10.1109/CVPR.2019.00363}
\showDOI{\tempurl}


\bibitem[\protect\citeauthoryear{Zheng, Song, Chen, Hu, Cao, and Nie}{Zheng
  et~al\mbox{.}}{2019}]%
        {Zheng2019ACMMM}
\bibfield{author}{\bibinfo{person}{Na Zheng}, \bibinfo{person}{Xuemeng Song},
  \bibinfo{person}{Zhaozheng Chen}, \bibinfo{person}{Linmei Hu},
  \bibinfo{person}{Da Cao}, {and} \bibinfo{person}{Liqiang Nie}.}
  \bibinfo{year}{2019}\natexlab{}.
\newblock \showarticletitle{Virtually Trying on New Clothing with Arbitrary
  Poses}. In \bibinfo{booktitle}{\emph{Proceedings of the ACM International
  Conference on Multimedia}}. \bibinfo{pages}{266–274}.
\newblock
\urldef\tempurl%
\url{https://doi.org/10.1145/3343031.3350946}
\showDOI{\tempurl}


\end{thebibliography}

\section*{Supplementary Material}
\appendix
\section{abstract}
  This supplementary material presents the detailed network structure, parameter settings, and training losses of ARShoe.
\section{Details of network architecture}
\begin{figure}[h]
	\centering	
	\includegraphics[width=0.9\linewidth]{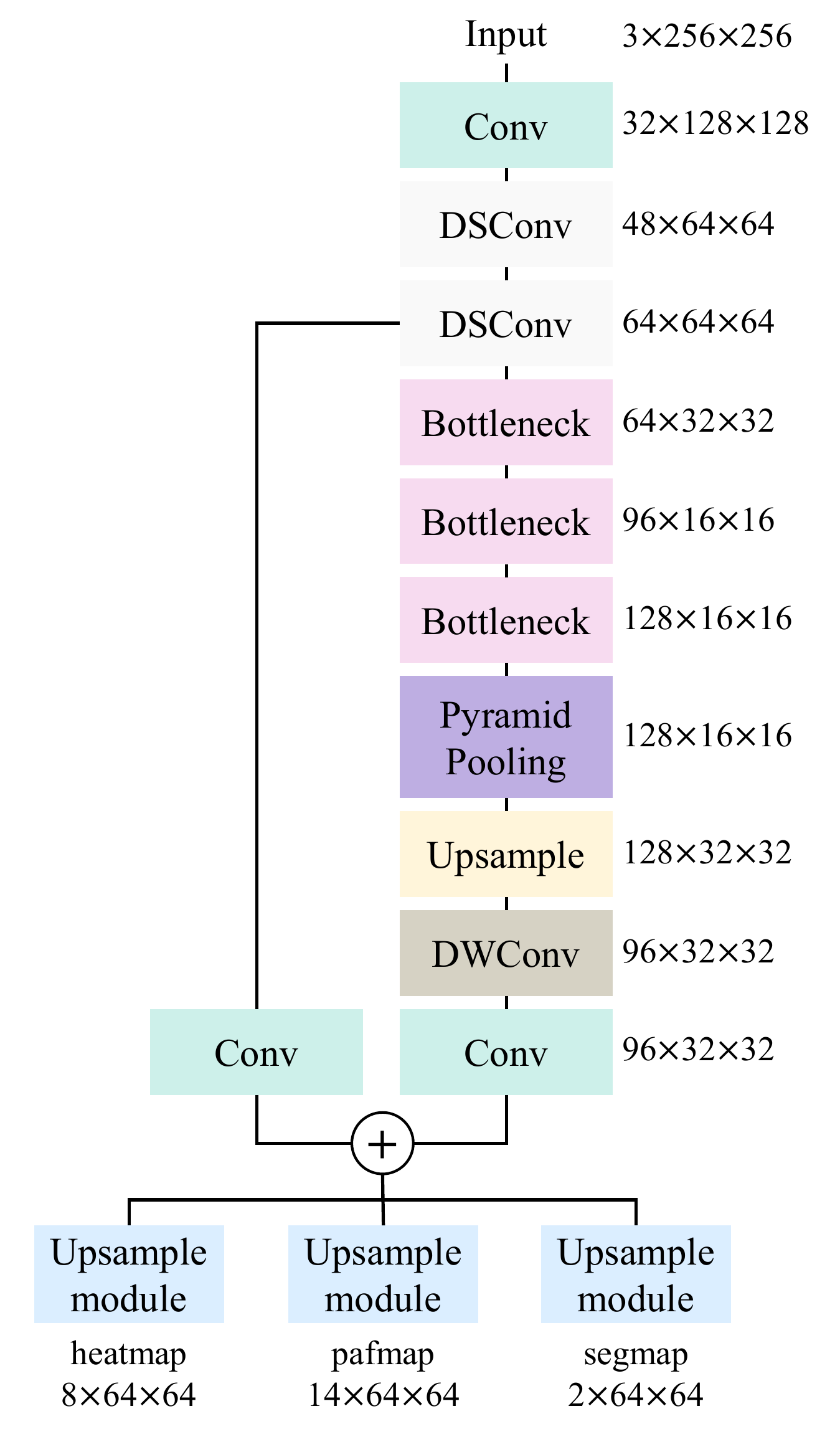}
	\caption{The detailed network architecture of ARShoe. 
	}
	\label{fig:supp}
\end{figure}

Figure~\ref{fig:supp} illustrates the detailed network architecture and parameter settings of our multi-branch network. Firstly, the encoder part extracts low-level features utilizing a convolution block and two depthwise separable convolution blocks (DSConv). The high-level features are then obtained through three residual bottleneck blocks (Bottleneck), a pyramid pooling block, and a depthwise convolution block (DWConv). Afterward, both low-level and high-level features go through a convolution block to unify their channel numbers. To integrate high-level semantic features and low-level shallow features for better image representation, high-level and low-level features are summated and go through an activation function ReLU. In the decoding process, three upsample modules are adopted to decode the fused features to keypoint heatmap (heatmap), PAFs heatmap (pafmap), and segmentation results (segmap). The dimension variation process of feature maps is presented in Fig.~\ref{fig:supp}.

\section{Training losses}
We utilize $\mathcal{L}_2$ loss for all three subtasks in our network. The ground truths of the keypoints prediction branch, the PAFs estimation branch, and the segmentation branch are the annotated keypoints, connections between keypoints, and segmentation of foot and leg, respectively. Denoting losses of these three branches as $\mathcal{L}_{\rm heatmap}$, $\mathcal{L}_{\rm pafsmap}$, and $\mathcal{L}_{\rm segmap}$, respectively, the overall loss $\mathcal{L}$ of our network can be fomulated as:
\begin{equation}
 	\mathcal{L} = \lambda_1 \mathcal{L}_{\rm heatmap} + \lambda_2 \mathcal{L}_{\rm pafsmap} + \lambda_3 \mathcal{L}_{\rm segmap}
\end{equation}
where $\lambda_1$, $\lambda_2$, and $\lambda_3$ are coefficients to balance the contributions of each loss. In our implementation, they are set to $2$, $1$, and $0.1$, respectively.
%
%
%
%
%
%
%

\end{document}